\documentclass[letterpaper, 10 pt, journal, twoside]{IEEEtran}  %

\IEEEoverridecommandlockouts                              %

\newif\ifrenderfigures %
\renderfigurestrue

\usepackage{siunitx}
\usepackage{xcolor}
\definecolor{OIblack}{RGB}{0, 0, 0}
\definecolor{OIgreen}{RGB}{0, 158, 115}
\definecolor{OIblue}{RGB}{0, 114, 178}
\definecolor{OIlightblue}{RGB}{86, 180, 233}
\definecolor{OIyellow}{RGB}{240, 228, 66}
\definecolor{OIorange}{RGB}{230, 159, 0}
\definecolor{OIred}{RGB}{213, 94, 0}
\definecolor{OIpink}{RGB}{204, 121, 167}

\usepackage{amsmath, amssymb, latexsym}

\usepackage{dsfont}

\usepackage{bm} %

\usepackage{calc}

\usepackage{cite}

\usepackage{booktabs}

\usepackage[acronym]{glossaries}

\newacronym{rl}{RL}{Reinforcement Learning}
\newacronym{il}{IL}{Imitation Learning}

\newacronym{mpc}{MPC}{Model Predictive Control}
\newacronym[firstplural={Degrees of Freedom} (DoF)]{dof}{DoF}{Degree of Freedom}
\newacronym{dh}{DH}{Denavit-Hartenberg}

\newacronym{sofa}{SOFA}{Simulation Open Framework Architecture}
\newacronym{pbd}{PBD}{position based dynamics}
\newacronym{fem}{FEM}{finite element method}

\newacronym{ras}{RAS}{Robot-Assisted Surgery}

\newacronym{sdf}{SDF}{Signed Distance Function}
\newacronym{iou}{IoU}{Intersection Over Union}
\newacronym{hd}{HD}{Hausdorff Distance}
\newacronym{icp}{ICP}{Iterative Closest Point}
\newacronym[plural=ROIs]{roi}{ROI}{Region of Interest}
\newacronym{uwc}{UWC}{Uncertainty Weighted Centroid}
\newacronym{spwc}{SPWC}{Softmax Probability Weighted Centroid}
\newacronym{on}{ON}{Occupancy Network}
\newacronym{iss}{ISS}{Improved Sort Sample} %
\newacronym{mcd}{MCD}{Monte Carlo Dropout}
\newacronym{fps}{FPS}{Farthest Point Sampling}
\newacronym{ct}{CT}{Computed Tomography}
\newacronym{v2s}{V2S}{Volume2SurfaceCNN}

\newacronym[shortplural={\ensuremath{\text{MLP}_{\theta_1}}}, longplural={Multilayer Perceptron}]{mlp}{MLP}{Multilayer Perceptron}
\newcommand{\glsspooky}[1]{\glspl{#1}}

\newacronym{methodname}{LUDO}{}

\usepackage{tikz}
\usetikzlibrary{backgrounds}
\usetikzlibrary{fit}
\usetikzlibrary{spy}
\usetikzlibrary{positioning} %
\usetikzlibrary{calc}
\usetikzlibrary{patterns} %
\usetikzlibrary{quotes} %
\usetikzlibrary{calc}
\usetikzlibrary{decorations.markings}
\usetikzlibrary{matrix, positioning}
\usepackage{pgfplots}
\pgfplotsset{compat=1.16}
\usepgfplotslibrary{fillbetween}

\usetikzlibrary{fit}
\tikzset{
  fitting node/.style={
    inner sep=0pt,
    fill=none,
    draw=none,
    reset transform,
    fit={(\pgf@pathminx,\pgf@pathminy) (\pgf@pathmaxx,\pgf@pathmaxy)}
  },
  reset transform/.code={\pgftransformreset}
}

\usepackage{xspace}

\newcommand*{\ie}{\emph{i.e.}\@\xspace}

\newboolean{includeHighly}
\setboolean{includeHighly}{false} %
\newcommand{\highly}{\ifthenelse{\boolean{includeHighly}}{highly}{}}
\newcommand{\Highly}{\ifthenelse{\boolean{includeHighly}}{Highly}{}}

\usepackage{hyperref}
\hypersetup{%
    colorlinks=false,
    allbordercolors={1 1 1},
    hidelinks
}

\usepackage{cleveref}

\def\equationautorefname#1#2\null{%
  Eq.#1(#2\null)%
} %

\usetikzlibrary{decorations.markings}

\tikzset{outside/.style={
    postaction={
        decorate,
        decoration={
            markings,
            mark=at position \pgfdecoratedpathlength-0.1pt with {\arrow[cyan,line width=#1] {>}; },
            mark=between positions 0 and \pgfdecoratedpathlength-1.5pt step 0.1pt with {
                \pgfmathsetmacro\myval{multiply(divide(
                    \pgfkeysvalueof{/pgf/decoration/mark info/distance from start}, \pgfdecoratedpathlength),100)};
                \pgfsetfillcolor{cyan!\myval!blue};
                \pgfpathcircle{\pgfpointorigin}{#1};
                \pgfusepath{fill};}
}}}}
\tikzset{inside/.style={
    postaction={
        decorate,
        decoration={
            markings,
            mark=at position \pgfdecoratedpathlength-0.1pt with {\arrow[yellow,line width=#1] {>}; },
            mark=between positions 0 and \pgfdecoratedpathlength-1.5pt step 0.1pt with {
                \pgfmathsetmacro\myval{multiply(divide(
                    \pgfkeysvalueof{/pgf/decoration/mark info/distance from start}, \pgfdecoratedpathlength),100)};
                \pgfsetfillcolor{yellow!\myval!red};
                \pgfpathcircle{\pgfpointorigin}{#1};
                \pgfusepath{fill};}
}}}}
\usepackage{colortbl}
\def\cca#1{\cellcolor{black!#10}\ifnum #1>5\color{white}\fi{#1}} %
\usepackage{pgf}
\usepackage{collcell}
\usepackage[accsupp]{axessibility} 

\newcommand*{\MinNumber}{0.28}%
\newcommand*{\MaxNumber}{0.9}%

\newcommand{\ApplyGradient}[1]{%
  \pgfmathsetmacro{\PercentColor}{100.0*(#1-\MinNumber)/(\MaxNumber-\MinNumber)}%
  \edef\x{\noexpand\cellcolor{blue!\PercentColor!red}}\x\textcolor{white}{#1}%
}
\newcolumntype{R}{>{\collectcell\ApplyGradient}{r}<{\endcollectcell}}

\title{LUDO: Low-Latency Understanding of \Highly Deformable Objects using Point Cloud Occupancy Functions}

\author{Pit Henrich$^{1}$,  Franziska Mathis-Ullrich$^{1}$, and Paul Maria Scheikl$^{1,2}$%
\thanks{$^{1}$ Department Artificial Intelligence in Biomedical Engineering, Friedrich-Alexander-University Erlangen-Nürnberg, 91052 Erlangen, Germany. \newline {\small Corresponding author: \tt franziska.mathis-ullrich@fau.de}}%
\thanks{$^{2}$ Laboratory for Computational Sensing and Robotics, Johns Hopkins University, Baltimore, MD 21218 USA.}%

}

\begin{document}

\maketitle

\begin{abstract}
Accurately determining the shape of deformable objects and the location of their internal structures is crucial for medical tasks that require precise targeting, such as robotic biopsies.
We introduce LUDO, a method for accurate low-latency understanding of deformable objects.
LUDO reconstructs objects in their deformed state, including their internal structures, from a single-view point cloud observation in under 30 ms using occupancy networks.
LUDO provides uncertainty estimates for its predictions.
Additionally, it provides explainability by highlighting key features in its input observations.
Both uncertainty and explainability are important for safety-critical applications such as surgery.
We evaluate LUDO in real-world robotic experiments, achieving a success rate of 98.9\% for puncturing various regions of interest (ROIs) inside \highly deformable objects.
We compare LUDO to a popular baseline and show its superior ROI localization accuracy, training time, and memory requirements. 
LUDO demonstrates the potential to interact with deformable objects without the need for deformable registration methods.
\end{abstract}

\section{INTRODUCTION}
\label{sec:introduction}

    \ifrenderfigures

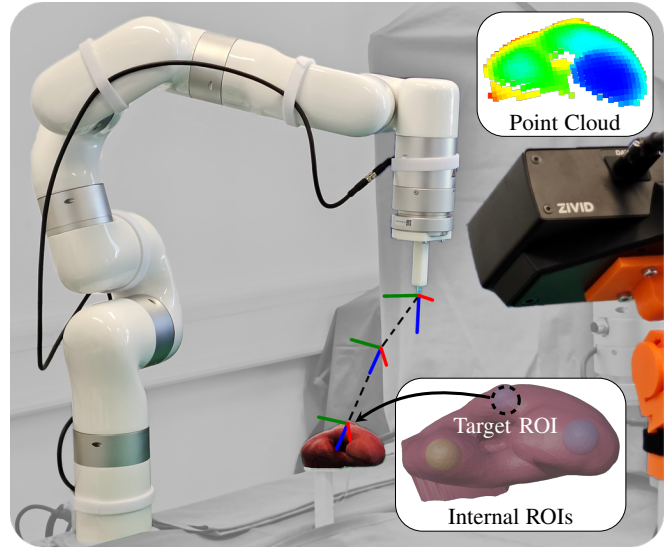
\begin{figure}[!ht]
    \centering
    \resizebox{1.0\columnwidth}{!}{
        \begin{tikzpicture}
        
        \node[anchor=south west, inner sep=0] (image) at (0,0) {\includegraphics[width=12cm]{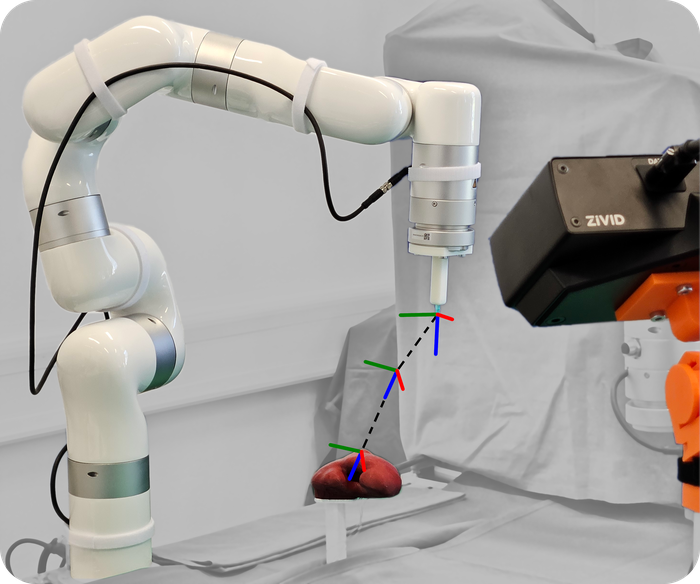}};

            \node[draw, rounded corners=0.5cm, fill=white, text width=4cm, align=center] 
    at (9.2,1.7) {\includegraphics[width=4cm]{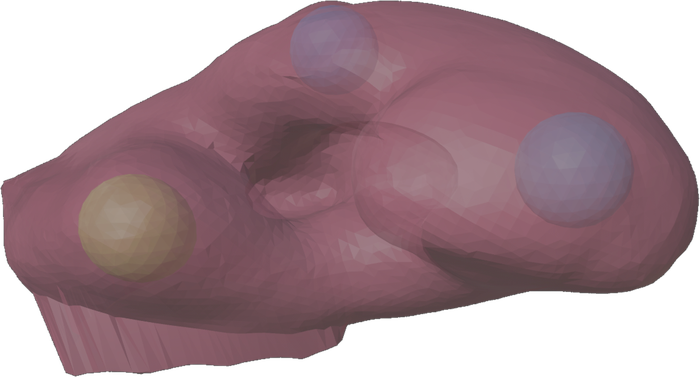}\\\large Internal ROIs};

            \node[draw, rounded corners=0.5cm, fill=white, text width=3cm, align=center] (PCD)
    at (10.2,8.7) {\includegraphics[width=3cm]{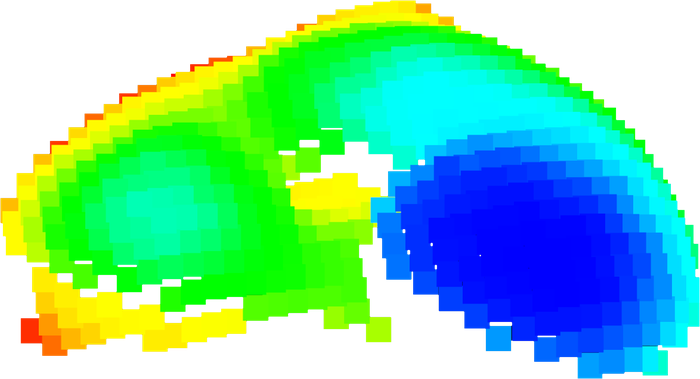}\\\large Point Cloud};

        \node (VROI) at (9.11,2.7) [draw, thick, dashed, circle,scale=1.5, ultra thick] {};
        \node (RROI) at (6.2,2.2) [circle,scale=1] {};
        \node (Zivid) at (10.2,6.5) [circle,scale=1.0] {};

        \node[inner sep=0, color=white, anchor=south] (TextROI) at (9.11,2.0) {\large Target ROI};
        \draw[-stealth, black, ultra thick] (VROI) to[out=170,in=35] (RROI);

        \end{tikzpicture}
    }
    \caption{
        Robotic setup used for real-world evaluation.
        An xArm7 robot (UFactory, China) automatically punctures Regions of Interests (ROIs) inside a deformable organ phantom using a needle instrument.
        The inset image displays the spherical \textit{Internal ROIs}, with the middle one marked as the \textit{Target ROI} for the needle puncture.
        The Zivid One+ M (Zivid, Norway) is used to capture a surface \textit{Point Cloud} of the phantom.
        This \textit{Point Cloud} is used by our proposed method to estimate the location of all ROIs in under 30 milliseconds, excluding depth image capture time.}
    \label{fig:setup_first_figure}
\end{figure}

    \fi

    \IEEEPARstart{I}{nferring} the shape of deformable objects and the location of their internal structures is important in medical procedures.
    Depending on the surgical intervention, different internal structures have to be either targeted or avoided.
    Examples include targeting tumorous tissue during biopsies and avoiding risk structures such as blood vessels and nerves.
    
    Preoperative imaging can provide shape information and the location of these internal structures before an intervention.
    However, during interventions, deformations cause deviations from the preoperative state.
    For example, deformations occur when the surgeon applies force to soft tissue by pushing or pulling.
    Moreover, preoperative imaging modalities are often not available intraoperatively, necessitating a way to determine the location and shape of internal structures from observation modalities that are available during the procedure.
    
    Addressing these challenges, we consider the following setting: Given the surface mesh of a deformable object in its non-deformed state, determine the location and shape of internal structures from a single-view point cloud observation of the object in its deformed state.
    In a medical scenario, the surface mesh can be derived from preoperative imaging such as segmented \gls{ct} scans.
    The point cloud, captured during an intervention, contains only partial information about the visible surface and typically lacks any features of the internal structures.

    \ifrenderfigures
        \input{figures/main_figure_v04_horizontal}
    \fi

    For this work, we consider a concrete robotic application: puncturing a \gls{roi} within a \highly deformable object using a needle, similar to a biopsy procedure.
    The robotic setup and puncturing task are shown in \Cref{fig:setup_first_figure}. 
    The task requires accurately localizing the \gls{roi} despite the object's deformation.
    Additionally, it is not enough to only localize the \glspl{roi} as the encompassing object's surface is needed to plan suitable trajectories.
    To minimize deformations caused by friction during the puncturing, the needle should traverse as little material as possible before reaching the \gls{roi}.
    
    To this end, we propose \gls{methodname}, a method that infers structural information of known deformable objects with low latency.
    Given a single-view point cloud of a deforming object's surface, \gls{methodname} reconstructs a dense point cloud that captures both the object’s shape and its internal structures.
    This is achieved using an \gls{on}, which provides an implicit 3D representation conditioned on the single-view point cloud.
    Instead of attempting to generalize across arbitrary shapes, \gls{methodname} learns deformation behaviors for specific prior objects.
    In a medical context, \gls{methodname} is therefore trained using patient-specific data derived from prior medical scans, such as \gls{ct} scans.
    Trained on data from deformation simulations, the \gls{on} can reconstruct the deformed object, including its internal structures during inference on real-world observations.
    The simulation requires only a single preoperative surface mesh, such as one obtained from a volume scan.
    For our robotic application, the reconstruction is then used to plan a puncture path to reach the \gls{roi}.
    The process of inferring and planning a trajectory based on a single-view point cloud is illustrated in \Cref{fig:process_overview}.

    Although learning-based deformable registration such as \gls{v2s}~\cite{pfeiffer2020non} already provide promising structural information about deforming objects, \gls{methodname} achieves higher \gls{roi} localization accuracy, trains faster, and requires orders-of-magnitude less storage.
    Further, \gls{methodname} does not require an initial alignment.
    
    The main contributions of this work are as follows:
    
    \begin{enumerate}
        \item \textbf{Low-Latency Structural Estimation}: We provide an accurate low-latency approach for obtaining structural information of \highly deformable objects using \glspl{on}, conditioned on single-view point cloud observations and trained on physics-based simulation data.
        \item \textbf{Robotic Puncturing Experiments}: We demonstrate the effectiveness of the proposed method through real-world robotic experiments for autonomous puncturing of targets within three \highly deformable phantoms (\ie, \textit{Organ}, \textit{Cylinder}, \textit{Slab}).
        \item \textbf{Uncertainty Estimation and Explainability}:
        We
            \begin{enumerate}
                \item integrate methods to compute uncertainty for deformable object understanding, 
                \item explore the use of probabilities and different uncertainty estimating methods for selecting an optimal puncture target position,
                \item conduct a preliminary exploration of estimating a single aggregated uncertainty value to assess whether data is reliable for downstream tasks, and
                \item integrate a method for explainability to highlight which key features in the input were used to determine the deformation.
            \end{enumerate}
        \item \textbf{Robot Calibration}: We provide practical considerations for robot calibration to improve the absolute accuracy of industrial robotic arms in high precision applications.
    \end{enumerate}
    
    By addressing the challenge of reconstructing structures of deformable objects from partial observations, \gls{methodname} is a potential step towards improving the safety and effectiveness of medical interventions that require accurate localization of internal anatomy.

    \section{RELATED WORK}
    \label{sec:related_work}
        \paragraph{Deformable Object Registration}
        \label{sec:related_work:dor}
            Surface-to-point-cloud registration is the process of aligning a surface mesh with a point cloud, commonly through rotations and translations.
            An optimal solution is found when the points in the point cloud are as close as possible to the mesh surface after registration.
            For rigid objects, methods such as \gls{icp}~\cite{besl1992method} and its derivatives are effective.
            However, there are many applications, such as medical tasks, where the registration target is deformable.
            A rigid registration approach will result in increasingly worse results as the amount of deformation increases.
            Deformable registration methods, in addition to rotations and translations, deform the surface mesh to better fit the point cloud observation.
            For example, this can be done by estimating an additional deformation field~\cite{pfeiffer2019learning, pfeiffer2020non}.
            This deformation field applies forces or translations to the vertices of the surface mesh to deform it.%

            Jia et al.~\cite{Jia_Kyan_2021} use neural occupancy functions to improve registration of human liver models.
            Instead of directly registering a preoperative surface mesh to the intraoperatively observed point cloud, they use the neural occupancy function conditioned on the point cloud as the registration target.
            Their neural occupancy function provides gradients for each point in 3D space, and this gradient field is used to deform the mesh for registration.
            Although their approach outperforms rigid registration methods, it is only viable for small deformations, as it requires a rigid alignment for initialization.
            Initial rigid alignment has inevitable ambiguities under strong deformations.
            The dependency on an initial alignment is common for optimization-based registration approaches, and often limits them to small deformations~\cite{deng2022survey}.

            In previous work~\cite{henrich2024registered}, we have shown that using \gls{on}-based reconstruction of deformable objects provides an alternative to deformable registration methods.
            Similar to Jia et al.~\cite{Jia_Kyan_2021}, an \gls{on} is conditioned on an input point cloud.
            However, in contrast to using the \gls{on} as a new registration target, we generate an object representation that can be used in place of a registered model.
            In addition, our approach does not require initialization through a rigid registration step and works for large deformations.
            
            In contrast to our previous work~\cite{henrich2024trackingtumorsdeformationpartialIEEE}, we utilize physics-based simulation to generate realistic deformations for our training data.
            Combined with a high-quality depth sensor, our proposed approach is able to accurately reconstruct objects in the real world.
            To address transparency and safety in high-risk applications such as surgery, we extend the method to provide initial quantitative measures of uncertainty and explainability.
            Finally, we demonstrate our method's effectiveness through real-world robotic experiments that require accurate reconstruction of internal structures in \highly deformable objects.

        \paragraph{Deformable Object Manipulation}
        \label{sec:related_work:dom}
            Interaction with deformable objects is investigated through classical and data-driven model-based methods.
            These approaches manipulate deformable objects and indirectly achieve the desired shapes of the internal structures for biopsies~\cite{afshar2022modelbasedmultipoint},  cryoablation~\cite{alambeigi2018semiautonomouscryoablation}, and suturing~\cite{zhong2019dualarmrobotic}.
            However, these approaches assume that the position and shape of the internal structures can be determined intraoperatively from sensor observations, for example through an ultrasound probe.
            This limits their applicability and further requires complex data processing to locate, segment, and reconstruct the internal structures from the ultrasound images.
            
            Florence et al.~\cite{florence2018dense} learn dense visual descriptors capturing correspondences across different object instances and views.
            This enables robots to manipulate deformable objects by understanding their visual geometry without explicitly modeling deformations.
            However, they do not model the internal structures of deformable objects.
            
            Other works investigate deformable object manipulation with the goal of reaching desired deformation states~\cite{thach2022learningvisual, ou2023simtorealsurgical, scheikl2024mpd}.
            However, they are also unable to reason about the deformation state of internal structures, which is essential for medical tasks such as biopsies.

\section{Method Overview}
\label{sec:methods}
    This work consists of two primary components: \gls{methodname} itself and the robotic experimental method.

    \Cref{sec:methods_reconstruction} details our proposed method, \gls{methodname}, for low-latency understanding of deformable objects using point cloud occupancy functions.
    \gls{methodname} estimates both the external and internal structures of deformable objects from single-view point cloud observations.
    
    \Cref{sec:methods_autonomous_puncturing} details our real-world robotic experiments.
    We describe the robotic system that uses the \gls{methodname}-generated output for targeting and puncturing tasks in \highly deformable objects.
    This includes an approach to improve the absolute accuracy of industrial robotic systems through \gls{dh} parameter optimization.

\section{LUDO} %
\label{sec:methods_reconstruction}
    \gls{methodname} provides 3D structural information of deformable objects in the form of a dense output point cloud, see \Cref{fig:process_overview}.
    \gls{methodname} uses an \gls{on} trained on realistic deformation data generated through \gls{fem} simulations.

    In this section, we first describe how known deformable objects can be reconstructed by a trained \gls{on}, conditioned on an observation point cloud.
    We detail how to produce training data for such \gls{on}, using a sampling strategy optimized for occupancy learning and a physics based deformation simulation.
    We describe techniques to provide local uncertainty values for every part of the inferred structures.
    Additionally, by aggregating the local uncertainty values we can provide a single aggregated uncertainty value.
    Finally, we provide a masking based explainability approach to identify features in the input point cloud that are decisive during inference.

    \subsection{Object Reconstruction with Occupancy Networks}
        \label{sec:methods_reconstruction_using_occupancy_networks}
    \ifrenderfigures
        \definecolor{OIblack}{RGB}{0, 0, 0}
\definecolor{OIgreen}{RGB}{0, 158, 115}
\definecolor{OIblue}{RGB}{0, 114, 178}
\definecolor{OIlightblue}{RGB}{86, 180, 233}
\definecolor{OIyellow}{RGB}{240, 228, 66}
\definecolor{OIorange}{RGB}{230, 159, 0}
\definecolor{OIred}{RGB}{213, 94, 0}
\definecolor{OIpink}{RGB}{204, 121, 167}

\def\myyslant{0.15}
\def\myxslant{0.0}
\def\mynoise{0.15}

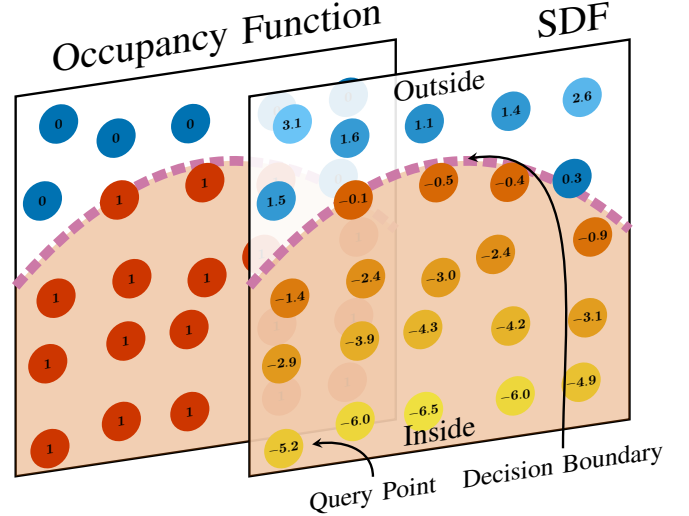
\begin{figure}
    \centering
    \resizebox{1.0\columnwidth}{!}{
        \begin{tikzpicture}[scale=1.7,every node/.style={minimum size=1cm},on grid]
        		
            \begin{scope}[
                    yshift=0,every node/.append style={
                    yslant=\myyslant,xslant=\myxslant},yslant=\myyslant,xslant=\myxslant
                    ]
                \fill[white,fill opacity=1.0] (0,0) rectangle (4,4);
                \draw[black,very thick] (0,0) rectangle (4,4);%
                parabola curve
                \fill [OIred, opacity=0.3] (0,2) parabola bend (2,3) (4,2) -- (4,0) -- (0,0) -- cycle;
                \draw [color=OIpink, line width=5pt, dash pattern=on 7pt off 3pt](0,2) parabola bend (2,3) (4,2)  ; %

                \node (UpperRight) at (4,3.7) [scale=2, anchor=south east] {Occupancy Function};

                \pgfkeys{/pgf/number format/.cd, fixed, zerofill, precision =0} 
        
                \pgfmathsetseed{42}
        
                \foreach \x in {0,...,4} {
                    \foreach \y in {0,...,4} {
                        \pgfmathparse{0.4 + 0.8*\x + (rand*\mynoise - \mynoise/2)}
                        \pgfmathresult \let\xpoint\pgfmathresult;
                        \pgfmathparse{0.4 + 0.8*\y + (rand*\mynoise - \mynoise/2)}
                        \pgfmathresult \global\let\ypoint\pgfmathresult;
                
                        \pgfmathparse{\xpoint}
                        \pgfmathresult \global\let\xx\pgfmathresult;
                          \foreach \iter in {1,...,4} {
                              \pgfmathparse{0.25*(\xx*\xx*\xx-6*\xx*\xx+4*(\xx-2)*\ypoint
                               +12*\xx-8*\xpoint+8)}
                              \pgfmathresult \let\functionderv\pgfmathresult;
                              \pgfmathparse{3*(\xx-2)*(\xx-2)/4+\ypoint}
                              \pgfmathresult \let\functiondervv\pgfmathresult;
                              \pgfmathparse{\xpoint-\functionderv/(\functiondervv)}
                             \pgfmathresult \let\xx\pgfmathresult;
                          }
                          
                          \pgfmathparse{-\xx*\xx/4+\xx+2}
                          \pgfmathresult \global\let\yy\pgfmathresult;
                          \pgfmathsetmacro{\dd}{sqrt((\xpoint-\xx)* (\xpoint-\xx)
                            + (\ypoint-\yy)*(\ypoint-\yy ))/.4};
                          \pgfmathparse{int(\yy*100)}
                          \pgfmathresult \let\yyy\pgfmathresult;
                          \pgfmathparse{int(\ypoint*100)}
                          \pgfmathresult \let\yypoint\pgfmathresult;
                          \ifnum \yyy > \yypoint { %
                              \pgfmathparse{-\dd} \pgfmathresult \global\let\dd\pgfmathresult;
                              }
                          \fi	
                
                          \ifnum \yyy > \yypoint { %
                              \pgfmathparse{1} \pgfmathresult \global\let\dd\pgfmathresult;
                              }
                          \fi	
                          \ifnum \yyy < \yypoint { %
                              \pgfmathparse{0} \pgfmathresult \global\let\dd\pgfmathresult;
                              }
                          \fi	
                
                        \ifnum \yyy > \yypoint { %
                             \pgfmathparse{-\dd * 30}
                             \pgfmathresult \let\percentageDistance\pgfmathresult;
    
                             \node at (\xpoint,\ypoint) [fill=OIyellow!\percentageDistance!OIred,circle,scale=0.7] {};
                             }
                         \fi
                        \ifnum \yyy < \yypoint { %
                             \pgfmathparse{\dd * 30}
                             \pgfmathresult \let\percentageDistance\pgfmathresult;
    
                             \node at (\xpoint,\ypoint) [fill=OIlightblue!\percentageDistance!OIblue,circle,scale=0.7] {};
                             }
                         \fi
                        \node[scale=0.7] at (\xpoint,\ypoint - 0.0)
                            {\textcolor{black}{$\mathbf{\pgfmathprintnumber{\dd}}$}};
                     }
                  }
        
            \end{scope}

            \begin{scope}[
                    xshift=70,every node/.append style={
                    yslant=\myyslant,xslant=\myxslant},yslant=\myyslant,xslant=\myxslant
                    ]
                \fill[white,fill opacity=0.9] (0,0) rectangle (4,4);
                \draw[black,very thick] (0,0) rectangle (4,4);%
                \fill [OIred, opacity=0.3] (0,2) parabola bend (2,3) (4,2) -- (4,0) -- (0,0) -- cycle;
                \draw [color=OIpink, line width=5pt, dash pattern=on 7pt off 3pt](0,2) parabola bend (2,3) (4,2)  ; %

                \node (UpperRight) at (4,3.7) [scale=2, anchor=south east] {SDF};
                \node (DecisionBoundary) at (2,3) {};

                \pgfkeys{/pgf/number format/.cd, fixed, zerofill, precision =1} 
                \pgfmathsetseed{42}
                \foreach \x in {0,...,4} {
                    \foreach \y in {0,...,4} {
                        \pgfmathparse{0.4 + 0.8*\x + (rand*\mynoise - \mynoise/2)}
                        \pgfmathresult \let\xpoint\pgfmathresult;
                        \pgfmathparse{0.4 + 0.8*\y + (rand*\mynoise - \mynoise/2)}
                        \pgfmathresult \global\let\ypoint\pgfmathresult;
                
                        \pgfmathparse{\xpoint}
                        \pgfmathresult \global\let\xx\pgfmathresult;
                          \foreach \iter in {1,...,4} {
                              \pgfmathparse{0.25*(\xx*\xx*\xx-6*\xx*\xx+4*(\xx-2)*\ypoint
                               +12*\xx-8*\xpoint+8)}
                              \pgfmathresult \let\functionderv\pgfmathresult;
                              \pgfmathparse{3*(\xx-2)*(\xx-2)/4+\ypoint}
                              \pgfmathresult \let\functiondervv\pgfmathresult;
                              \pgfmathparse{\xpoint-\functionderv/(\functiondervv)}
                             \pgfmathresult \let\xx\pgfmathresult;
                          }
                          
                          \pgfmathparse{-\xx*\xx/4+\xx+2}
                          \pgfmathresult \global\let\yy\pgfmathresult;
                          \pgfmathsetmacro{\dd}{sqrt((\xpoint-\xx)* (\xpoint-\xx)
                            + (\ypoint-\yy)*(\ypoint-\yy ))/.4};
                          \pgfmathparse{int(\yy*100)}
                          \pgfmathresult \let\yyy\pgfmathresult;
                          \pgfmathparse{int(\ypoint*100)}
                          \pgfmathresult \let\yypoint\pgfmathresult;
                          \ifnum \yyy > \yypoint { %
                              \pgfmathparse{-\dd} \pgfmathresult \global\let\dd\pgfmathresult;
                              }
                          \fi

                        \ifnum \yyy > \yypoint { %
                             \pgfmathparse{-\dd * 15}
                             \pgfmathresult \let\percentageDistance\pgfmathresult;
    
                             \node at (\xpoint,\ypoint) [fill=OIyellow!\percentageDistance!OIred,circle,scale=0.7] {};
                             }
                         \fi
                        \ifnum \yyy < \yypoint { %
                             \pgfmathparse{\dd * 40}
                             \pgfmathresult \let\percentageDistance\pgfmathresult;
    
                             \node at (\xpoint,\ypoint) [fill=OIlightblue!\percentageDistance!OIblue,circle,scale=0.7] {};
                             }
                         \fi
                        \node[scale=0.7] at (\xpoint,\ypoint - 0.0)
                            {\textcolor{black}{$\mathbf{\pgfmathprintnumber{\dd}}$}};
                     }
                  }
        
            \pgfmathsetseed{42}
            \pgfmathparse{0.4 + 0.8*0 + (rand*\mynoise - \mynoise/2)}
            \pgfmathresult \let\xpoint\pgfmathresult;
            \pgfmathparse{0.4 + 0.8*0 + (rand*\mynoise - \mynoise/2)}
            \pgfmathresult \global\let\ypoint\pgfmathresult;
            \node (QueryPoint) at (\xpoint,\ypoint) {};
        
            \draw[-stealth,very thick](3.3,-0.2) node[scale=1.3, anchor=north, inner sep=0, outer sep=-7pt]{Decision Boundary}
                 to[out=90,in=0] (DecisionBoundary);
        
            \draw[-stealth, very thick](1.3,-0.2) node[scale=1.3, anchor=north, inner sep=0, outer sep=-7pt]{Query Point} to[out=90,in=0] (QueryPoint);

            \node at (2,0.18) [scale=1.5, color=black] {Inside};
            \node at (2,4-0.18) [scale=1.5, color=black] {Outside};

            \end{scope}

        \end{tikzpicture}
    }

    \caption{Visualization of an \textit{Occupancy Function} and a \textit{Signed Distance Function (SDF)}. The \textit{Occupancy Function} only encodes inside and outside of an object using binary values of 1 and 0 respectively. The SDF provides the distance to the \textit{Decision Boundary} and uses the sign (+ or -) to indicate if the point is inside or outside the object.}
    \label{fig:occupancy_vs_sdf}
\end{figure}
    \fi

        An occupancy function
        \begin{equation}
            f: q \rightarrow s, \quad q \in \mathbb{R}^3, s \in \mathbb{N}_0
        \end{equation}
        represents a 3D object, consisting of multiple segments, by mapping Cartesian query points $q$ to scalar values $s$.
        Each value of $s$ represents a different segment of the object.
        Clusters of points with the same value $s$ form the segments.
        For example, an organ model with a single internal tumor may be represented using $s\in\{0,1,2\}$, where points with the values $1$ or $2$ are inside the organ or tumor, respectively.
        The value of $0$ is reserved for all points outside of the object.
        A visual example of an occupancy function is shown in \Cref{fig:occupancy_vs_sdf}.

        The occupancy function $f$ can also be conditioned on an observation $o$ so that 
            \begin{equation}
                f: q, o \rightarrow s, \quad q \in \mathbb{R}^3, o \in \mathbb{R}^m, s \in \mathbb{N}_0
            \end{equation}
        can represent different objects, different deformation states, or different object poses.
        Without the observation, the occupancy function can only represent a single object in a single state.
            
        An \gls{on} $\mathcal{M}_\theta(\cdot)$ can be used to approximate $f$, where $\theta$ represents learnable parameters.
        The learnable parameters are optimized by minimizing the term:
            \begin{equation}
                \underset{\theta}{\text{min}} \sum_{i} \sum_{j} \mathcal{L}^c\big(f(q_j, o_i), \mathcal{M}_\theta(q_j,o_i)\big),
            \end{equation}
        \noindent where $i$ and $j$ iterate over the observations and query points, respectively.  
        We use a cross entropy loss $\mathcal{L}^c$:
            \begin{equation}
                \mathcal{L}^c =
                \sum_s -\mathds{1}(f(q, o) = s) \log\big(\mathcal{M}_\theta(s|q,o)\big),
                \label{eq:optimization_problem_occupancy_only}
            \end{equation}
        where $\mathds{1}(\cdot)$ is $1$ if the query point $q$ belongs to segment $s$ and $0$ otherwise, and
        $\mathcal{M}_\theta(s|q,o)$ is the probability that $q$ belongs to the segment $s$ given the observation $o$.
        We estimate $\mathcal{M}_\theta(s|q,o)$ by applying the softmax function to the model output logits.

        The ground truth values from $f$ do not provide a signal that indicates how close the estimate of $\mathcal{M}_\theta$ is to the correct label.
        Points very far from the object of interest should provide a strong training signal if they are incorrectly classified as being inside of the object.
        This signal is important to instill the concept of the compactness of objects in $\mathcal{M}_\theta$ (an object being confined to a specific area).
        It prevents the object from extending infinitely or from having clusters appearing far away from the main object.
        \input{figures/l1_loss_for_occupancy_learning}
        The signal can be provided by an additional loss based on the query point distance to the nearest surface~\cite{henrich2024registered}.
        A qualitative example for the effect of using an additional distance loss is shown in \Cref{fig:advantage_of_l1}.
        A \gls{sdf}
            \begin{equation}
                d:q,o \rightarrow r^{\pm}, \quad q\in\mathbb{R}^3, o\in\mathbb{R}^m, r^{\pm}\in\mathbb{R}
            \end{equation}
        provides the distance $r^{\pm}$ of a query point $q$ to the nearest object surface.
        
        The sign of the distance indicates if the point is inside or outside of the object, see \Cref{fig:occupancy_vs_sdf}.

        $\mathcal{M}_\theta(\cdot)$ can be trained to approximate $(f \times d) : \mathbb{R}^3 \times \mathbb{R}^m \rightarrow \mathbb{N}_0 \times \mathbb{R}$ to model both occupancy and signed distance functions.
            
        For training, we use an L1 loss for the signed distance
            \begin{equation}
                \mathcal{L}^\pm = \left| d(q,o) - \mathcal{M}_\theta(q,o)\right|.
            \end{equation}
        For readability, we assume that $\mathcal{L}^c$ uses the class and $\mathcal{L}^\pm$ uses the signed distance output of $\mathcal{M}_\theta$.

        We combine $\mathcal{L}^c$ and $\mathcal{L}^\pm$ into the final loss function

        \begin{equation}
            \mathcal{L}_{\lambda} = \mathcal{L}^c
            + \lambda\mathcal{L}^\pm, 
        \end{equation}
        where $\lambda$ is a weighting factor to balance the loss components.

        The final optimization problem is
            \begin{equation}
                \underset{\theta}{\text{min}} \sum_{i} \sum_{j} \mathcal{L}_{\lambda}(q_j,o_i).
                \label{eq:optimization_problem}
            \end{equation}

        Using a point cloud $P = \{p_0, p_1, \cdots\} \subset \mathbb{R}^3$ as the conditioning observation requires additional considerations.
        The naive approach of concatenating all points into an observation vector $o = (p_0^\intercal, p_1^\intercal, \cdots) \in \mathbb{R}^m$ results in an order dependent representation, where $(p_0^\intercal, p_1^\intercal, \cdots) \neq (p_1^\intercal, p_0^\intercal, \cdots)$.

        Point cloud encoders address the issue by distilling a point cloud into a latent vector in an order-invariant manner, often utilizing commutative operations.
        We therefore use a point cloud encoder $\text{E}_{\theta_2}$ to generate a latent representation of the point cloud observation $P$, and use this latent representation as conditioning $o$ for the point cloud \gls{on}:
        \begin{equation}
            \mathcal{M}_\theta(q,o) = \mathcal{M}^\text{E}_\theta(q,P) = \text{MLP}_{\theta_1}(q,\underbrace{\text{E}_{\theta_2}(P)}_o),
        \end{equation}
        where $\text{E}_{\theta_2}$ encodes the point cloud $P$ and the \glsspooky{mlp} estimates the occupancy of the query point $q$ with respect to the encoding.
        The learnable parameters are $\theta = (\theta_1,\theta_2)$.

        In previous work~\cite{henrich2024registered}, we have shown the suitability of PointNet++~\cite{qi2017pointnet++} as the point cloud encoder $\text{E}_{\theta_2}$ for 3D reconstruction tasks using conditioned \glspl{on}.
        A fully connected \gls{mlp} has been shown to be suitable for conditionally mapping query points to segment labels and distances~\cite{mescheder2019occupancy, park2019deepsdf}.

        The density of the input point cloud depends on the sensor's distance during data acquisition.
        Closer depth sensors capture more surface points, while farther ones capture fewer.
        Unlike synthetic data, real-world point clouds often contain holes where depth data could not be estimated.
        To ensure that the point cloud encoder is robust to varying point cloud densities and incomplete data, we randomly drop points from the input point cloud $P$ during training.
        For our application, we found that dropping $50\%$ of input points results in no degradation of \gls{methodname}'s performance.

        We only use the positional data for $P$.
        Incorporating color in the point cloud would require accurate surface textures during training.
        Prior models, such as ones derived from CT scans, only provide structural information.
        Additionally, textures complicate the sim-to-real transfer, as the training data would have to be rendered realistically or extensive visual augmentation would have to be used.

        Neural networks tend to learn smoothed 3D representations, missing high frequency details~\cite{pmlr-v97-rahaman19a, mildenhall2020nerf}.
        To increase the sensitivity of $\mathcal{M}^\text{E}_\theta$ to small positional changes in the query points, we use our previously introduced \textit{negative-exponent} sinusoidal encoding~\cite{henrich2024registered}
        \begin{multline}
            \beta(q) = \left( \sin( 2^{-4} \pi q ), \cos( 2^{-4} \pi q), ..., \right. \\
                \left. \sin( 2^{5} \pi q), \cos( 2^{5} \pi q) \right)
        \end{multline}
        to encode all query points $q \in Q \subset \mathbb{R}^3$.
        Both $\sin$ and $\cos$ are applied element-wise.
        Note the addition of the negative exponents which is not used in work such as NeRF~\cite{mildenhall2020nerf}.
        
        \begin{figure}
    \centering
    
    \resizebox{1.0\columnwidth}{!}{
        \begin{tikzpicture}[scale=1.8,line cap=round,line join=round]

            \definecolor{OIblack}{RGB}{0, 0, 0}
            \definecolor{OIgreen}{RGB}{0, 158, 115}
            \definecolor{OIblue}{RGB}{0, 114, 178}
            \definecolor{OIlightblue}{RGB}{86, 180, 233}
            \definecolor{OIyellow}{RGB}{240, 228, 66}
            \definecolor{OIorange}{RGB}{230, 159, 0}
            \definecolor{OIred}{RGB}{213, 94, 0}
            \definecolor{OIpink}{RGB}{204, 121, 167}
            
            \coordinate (camera) at (-3,0);
            
            \draw[thick]
              (camera) ++(0,0) coordinate (C)
              -- ++(0.4,-0.2)
              -- ++(0,0.4)
              -- cycle; %
            \draw[thick]
              (camera) ++(0.4,0.2) rectangle ++(0.3,-0.4); %
            
            \draw[dashed] (camera) -- (0.1,-1);
            \draw[dashed] (camera) -- (-0.1,1);
            
            \node[above] at ($(camera)!0.45!(0,-0.5)$) {\parbox{1.5cm}{\centering View Frustum}};
            \node[above] at ($(camera)!0.99!(-2.7,0.5)$) {Depth Sensor};
            \node[above] at (2,1.) {Object};

            \begin{scope}[shift={(0.8,0)}] %
            
                \draw[opacity=0.3, fill=OIred,draw=none]
                  plot [smooth cycle] coordinates {
                    (1.3,0.8) (1.2,0) (1.1,-0.7) (0,-1.2)
                    (-1.1,-0.7) (-1.3,0) (-1.2,0.8) (0,1.3)
                  };

                \draw[OIpink,thick]
                  plot [smooth cycle] coordinates {
                    (1.3,0.8) (1.2,0) (1.1,-0.7) (0,-1.2)
                    (-1.1,-0.7) (-1.3,0) (-1.2,0.8) (0,1.3)
                  };
            
                \begin{scope}[shift={(-0.02,0)}]
                  \clip (-2,-1) rectangle (0,1);
                  \draw[dashed, ultra thick, OIgreen]
                    plot [smooth cycle] coordinates {
                      (1.3,0.8) (1.2,0) (1.1,-0.7) (0,-1.2)
                      (-1.1,-0.7) (-1.3,0) (-1.2,0.8) (0,1.3)
                    };
                \end{scope}
            
                \node[above, OIgreen, anchor=west] at (-1.3,-.4) {\parbox{2cm}{\centering Point Cloud Observation $P$}};
                  
            \end{scope}

            \draw[OIblue, fill=OIblue, opacity=0.00,draw=none] (-1,-1) rectangle (1,1);
            \draw[OIblue] (-1,-1) rectangle (1,1);
            
            \node[OIblue, anchor=west] at (-0.1,0.6) {\parbox{2cm}{Normalized Space}};
            
            \node[OIblue,anchor=south west] at (-1,1) {$(-1,1)$};
            \node[OIblue,anchor=south east] at (1,1) {$(1,1)$};
            \node[OIblue,anchor=north west] at (-1,-1) {$(-1,-1)$};
            \node[OIblue,anchor=north east] at (1,-1) {$(1,-1)$};
            \draw[fill=OIblue, draw=none] (-1,1) circle (0.04);
            \draw[fill=OIblue, draw=none] (1,1) circle (0.04);
            \draw[fill=OIblue, draw=none] (-1,-1) circle (0.04);
            \draw[fill=OIblue, draw=none] (1,-1) circle (0.04);

        \end{tikzpicture}
    }
    \caption{2D example of a scene containing a \textit{Depth Sensor} and an \textit{Object}. The depth sensor produces a \textit{Point Cloud Observation} $P$ within its \textit{View Frustum}. This observation $P$ is normalized to $[-1,1]^2$. As $P$ is only a partial observation, the normalized range will not contain the full \textit{Object}. To reason about regions outside the normalization space, a neural occupancy function must use a positional encoding that can uniquely represent points outside the normalization range.}
    \label{fig:importance_of_negative_exp_po}
\end{figure}

        Because sinusoidal functions are periodic, all queries are performed within a normalized space to ensure that each query point can be uniquely encoded.
        The observation point cloud $P$ reflects the real-world scale of the captured objects.
        Since the labeling of $Q$ is done in the coordinates of the point cloud $P$, we must normalize $P$.
        We calculate a uniform scaling factor across all dimensions and offset $P$ so that it lies within the range $[-1,1]^3$.
        Using negative exponents stretches the periods of the sinusoidal functions in our encoding, ensuring that points outside $[-1,1]^3$ can still be uniquely represented and queried.
        This is important because the normalization is based on partial observations $P$ of the object, making it necessary to handle queries outside this space to obtain a reconstruction of the full object, see \Cref{fig:importance_of_negative_exp_po}.

    \ifrenderfigures

\newdimen\OccupancyNetworkX
\newdimen\OccupancyPointCloudY

\newcommand{\textnode}[3][2cm]{%
    \node (#3) [draw, rectangle, align=center, #2] {%
        \shortstack{%
            \parbox{#1}{\centering #3}%
        }%
    };%
}

\newcommand{\textnodedown}[2]{
    \node (#2) [draw, rectangle,align=center, #1] {
        \shortstack{
            \parbox{2cm}{\centering \rotatebox{90}{#2}}
        }
    };
}

\newcommand{\textnodedownmath}[3]{
    \node (#2) [draw, rectangle,align=center, #1] {
        \shortstack{
            \parbox{2cm}{\centering \rotatebox{90}{#3}}
        }
    };
}

\newcommand{\textnodemath}[3]{
    \node (#2) [draw, rectangle,align=center, #1] {
        \centering #3
    };
}

\newcommand{\textnodemathnoborder}[3]{
    \node (#2) [rectangle,align=center, #1] {
        \centering #3
    };
}

\newcommand{\textnodedownmathtinybound}[3]{
    \node (#2) [draw, rectangle, align=center, #1] {
            \centering \rotatebox{90}{\tiny #3}
    };
}

\newcommand{\randompoints}[1]{
        \begin{tikzpicture}
            \pgfmathsetseed{3245} %
            \foreach \i in {1,...,30} {
                \pgfmathsetmacro{\x}{rand*0.4} %
                \pgfmathsetmacro{\y}{rand*0.3} %
                \node[fill=black, circle, inner sep=0.5pt] at (\x, \y) {};
            }
            \pgfmathsetmacro{\x}{rand*0.4} %
            \pgfmathsetmacro{\y}{rand*0.3} %
            \node[fill=#1, circle, inner sep=1.5pt] at (\x, \y) {};
        \end{tikzpicture}
}

\newcommand{\bigpoint}[1]{%
    \begin{tikzpicture}[baseline={(0,0)}]
        \node[fill=#1, circle, inner sep=1.5pt] (p) at (1.5pt, 0) {};
        \node[] (p) at (0, 0) {}; %
    \end{tikzpicture}%
}

\newcommand{\mlp}[1]{
        \begin{tikzpicture}[node distance=0.1cm]
                \node (Input) [draw, rectangle,align=center]{\centering \rotatebox{90}{\tiny Input}};
                \node (A) [draw, rectangle,align=center, right=of Input]{\centering \rotatebox{90}{\tiny 1024 + $b$}};
                \node (B) [draw, rectangle,align=center, right=of A]{\centering \rotatebox{90}{\tiny 512}};
                \node (C) [draw, rectangle,align=center, right=of B]{\centering \rotatebox{90}{\tiny 512}};
                \node (D) [draw, rectangle,align=center, right=of C]{\centering \rotatebox{90}{\tiny 512}};
                \node (E) [draw, rectangle,align=center, right=of D]{\centering \rotatebox{90}{\tiny 1024 + $b$ + 512}};
                \node (F) [draw, rectangle,align=center, right=of E]{\centering \rotatebox{90}{\tiny 512}};
                \node (G) [draw, rectangle,align=center, right=of F]{\centering \rotatebox{90}{\tiny 512}};
                \node (H) [draw, rectangle,align=center, right=of G]{\centering \rotatebox{90}{\tiny 512}};
                \node (I) [draw, rectangle,align=center, right=of H]{\centering \rotatebox{90}{\tiny $C$ + 1}};

                \draw[->] (Input) -- (A);
                \draw[->] (Input.south) to[bend right] (E.south);
                \draw[->] (A) -- (B);
                \draw[->] (B) -- (C);
                \draw[->] (C) -- (D);
                \draw[->] (D) -- (E);
                \draw[->] (E) -- (F);
                \draw[->] (F) -- (G);
                \draw[->] (G) -- (H);
                \draw[->] (H) -- (I);
        \end{tikzpicture}
}

\begin{figure}
    \centering
     \resizebox{0.9\columnwidth}{!}{
        \begin{tikzpicture}[node distance=0.2cm, auto]
          \begin{scope}[local bounding box=firstBox]
              
            \node (Sensor Point Cloud) [align=center] {
                \shortstack{
                    \includegraphics[width=0.8cm]{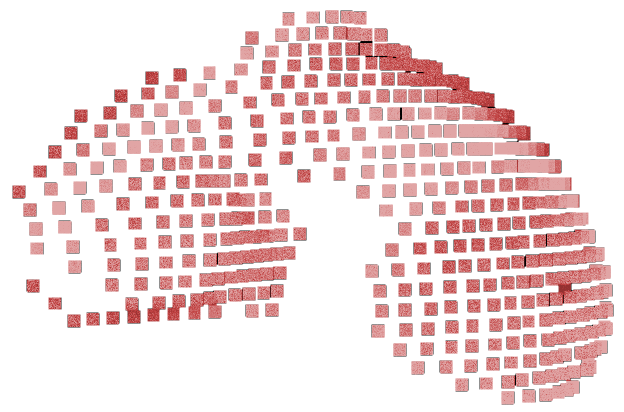}
                    \\
                    \parbox{2cm}{\centering $P$}
                }
            };

            \node (Query Point) [right=of Sensor Point Cloud, align=center] {
                {\shortstack{\randompoints{black}\\$Q \ni q = (x,y,z)$}}
            };

            \textnode[4cm]{below=of Sensor Point Cloud, xshift=1cm, yshift=-0.2cm}{Normalization}
            \textnode{below=of Normalization, xshift=0.5cm, yshift=-0.5cm}{Point Drop}
            \textnode{below=of Point Drop, yshift=-0.5cm}{PointNet++}
            
            \textnodemathnoborder{below=of PointNet++, yshift=-0.5cm}{Latent}{$(l_1,\cdots,l_{1024}) \times \beta(q) \in \mathbb{R}^{1024\times b}$}
            
            \textnodemath{below=of Latent, yshift=-0.5cm}{MLP}{\shortstack{\mlp\\\\MLP}}

            \textnodemathnoborder{below=0.4cm of MLP}{Output}{\shortstack{\bigpoint{magenta}\\\textcolor{magenta}{Occupancy} + Signed Distance}}

            \draw[->, thick, >=stealth] (Sensor Point Cloud.south) -| (Normalization) ;
            \draw[->, thick, >=stealth] (Normalization.south) -- ($(Point Drop.north) - (0.5cm, 0.0cm)$) node[midway, right, color=gray] {\scriptsize $P \leftarrow \text{Normalize}(P)$};
            \draw[->, thick, >=stealth] (Point Drop) -- (PointNet++) node[midway, right, color=gray] {\scriptsize $P\leftarrow \hat{P}, \hat{P}\subset P$};

            \draw[->, thick, >=stealth] (Query Point.south) -| node[pos=0.8, sloped, above] {\footnotesize{append}} ($(Latent.north east) - (0.3cm, 0.0cm)$);
            \draw[->, thick, >=stealth] (PointNet++) -- (Latent) node[midway, left, color=gray] {\scriptsize $l \leftarrow E^{\text{PointNet++}}(P)$};
            \draw[->, thick, >=stealth] (Latent) -- (MLP) node[midway, right, color=gray] {\scriptsize Input};
            \draw[->, thick, >=stealth] (MLP) -- (Output);

            \node (Container) [draw, black, fit=(Point Drop) (PointNet++) (Latent) (MLP) (Normalization), inner sep=0.15cm, dashed, thick,color=gray] {};

            \node (Occupancy Network) at ($(Container.north west)+(0,-1cm)$) [below, inner sep=2mm, anchor=north west]{ %
                \shortstack{
                    \includegraphics[height=0.8cm]{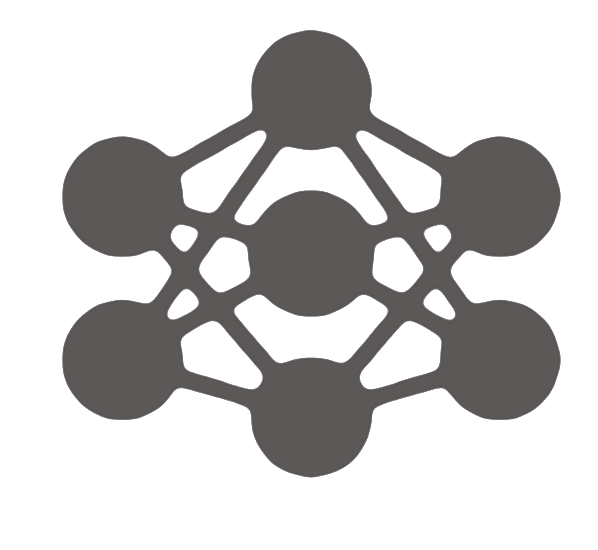}\\
                    \parbox{1cm}{\centering \tiny Occupancy Network}
                }
            };
        
          \end{scope}
        \end{tikzpicture}
     }
    \caption{
        The point cloud observation $P$, obtained from a depth sensor, is normalized to fit inside $[-1,1]^3$ (preserving aspect-ratios).
        Up to $50\%$ of points from $P$ are dropped randomly by Point Drop and the resulting point cloud is passed to the PointNet++ encoder to distill a latent representation $(l_1,\cdots,l_{1024})$.
        The query point, taken from $Q$, is encoded by $\beta$ and then appended to the latent.
        The concatenated vector is passed to the \textit{MLP}.
        Additionally, as skip connection from the input to the fifth layer is used.
        Between each layer, ReLU is used as an activation function.
        Batch normalization is used for the \textit{MLP}.
        The output is the label for the original query point and the signed distance.
    }
    \label{fig:architecture}
\end{figure}
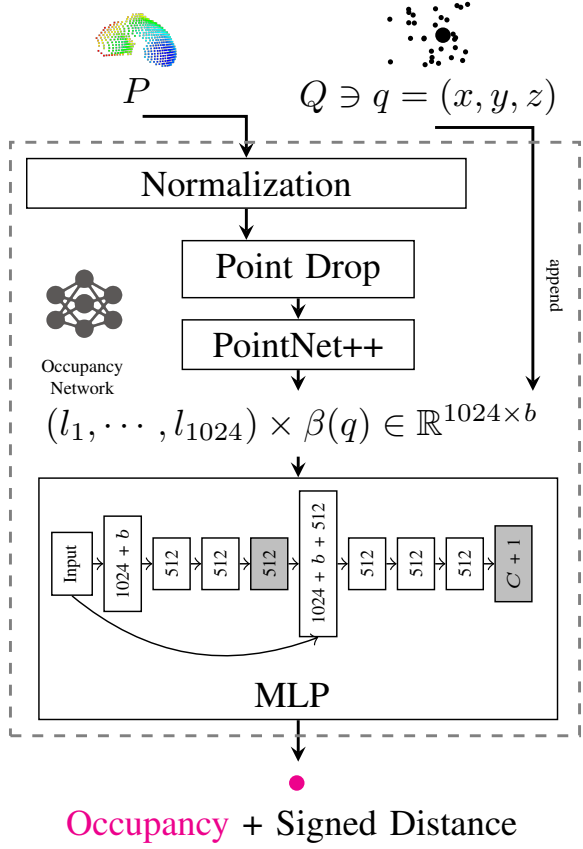

    \fi

        The architecture of \gls{methodname} is illustrated in \Cref{fig:architecture}.
        The used hyperparameters are listed in \Cref{tab:hyper}.

        \noindent\textbf{Ground Truth Occupancy Point Cloud:}
            \glspl{on} are trained through supervised learning.
            Each training sample is a tuple $(Q^\text{gt},P)$, where $Q^\text{gt} \subset \mathbb{R}^3 \times \mathbb{N}_0\times\mathbb{R}$ is a segment labelled point cloud with distances to the nearest surface, and $P$ is a point cloud observation.
            $Q^\text{gt}$, which we will refer to as an occupancy point cloud, contains ground truth values that are used in place of the function $(f \times d)$.
            The performance of the trained \gls{on} depends on how the points in $Q^\text{gt}$ are distributed~\cite{henrich2024registered}.
            Ideally, points in $Q^\text{gt}$ should be close to the surfaces between segments.
            Such points help the network to accurately define the boundaries between segments, leading to more precise part segmentation and better surface details.

    \ifrenderfigures
        \definecolor{OIblack}{RGB}{0, 0, 0}
\definecolor{OIgreen}{RGB}{0, 158, 115}
\definecolor{OIblue}{RGB}{0, 114, 178}
\definecolor{OIlightblue}{RGB}{86, 180, 233}
\definecolor{OIyellow}{RGB}{240, 228, 66}
\definecolor{OIorange}{RGB}{230, 159, 0}
\definecolor{OIred}{RGB}{213, 94, 0}
\definecolor{OIpink}{RGB}{204, 121, 167}

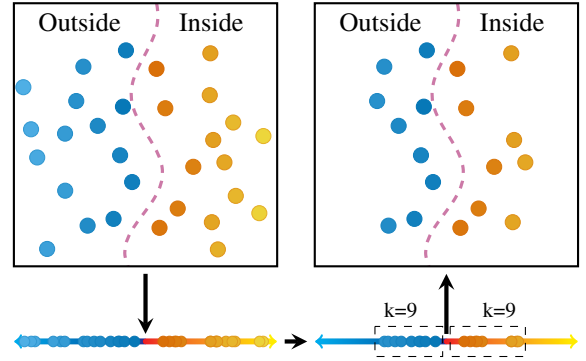
\begin{figure}[tb]
    
    \parbox{\columnwidth}{
        \centering
        \begin{tikzpicture}[node distance=0.1]
            \draw[domain=8:4.5, smooth, variable=\x, OIpink, very thick, dashed] plot ({1.3*sin(deg(\x*3))/-\x + 1.75}, {\x});
            \draw[draw=black, thick] (0,8) rectangle (3.5, 4.5);
            \node[anchor=north] at (0.875, 8) {Outside};
            \node[anchor=north] at (2.625, 8) {Inside};

            \draw[domain=8:4.5, smooth, variable=\x, OIpink, very thick, dashed] plot ({1.3*sin(deg(\x*3))/(-\x) + 5.75}, {\x});
            \draw[draw=black, thick] (4,8) rectangle (7.5, 4.5);
            \node[anchor=north] at (4.875, 8) {Outside};
            \node[anchor=north] at (6.625, 8) {Inside};

            \path [outside=1pt] (1.75,3.5) .. controls (0.875, 3.5) .. (0.0,3.5);
            \path [inside=1pt] (1.75,3.5) .. controls (2.625, 3.5) .. (3.5,3.5);
            
            \path [outside=1pt] (5.75,3.5) .. controls (4.875, 3.5) .. (4.0,3.5);
            \path [inside=1pt] (5.75,3.5) .. controls (6.625, 3.5) .. (7.5,3.5);

            \draw [-stealth, ultra thick] (1.75, 4.4) -- (1.75, 3.6); %
            \draw [-stealth, ultra thick] (3.6, 3.5) -- (3.9, 3.5); %
            \draw [-stealth, ultra thick] (5.75, 3.6) -- (5.75, 4.4); %

            \def\points{
            0.2/6, 0.5/6, 0.8/6, 1.1/6, 1.4/6, 1.7/6, 2.0/6,
            1.6/7.5,
            }

            \def\pointsInside{
                3.317/6.233, 3.263/5.207, 2.916/6.413, 2.712/4.725, 2.950/5.437, 2.604/6.777, 2.632/5.087, 2.612/7.334, 2.799/5.883, 2.650/6.181, 2.013/6.603, 2.385/5.824, 2.178/5.269, 1.936/5.012, 1.896/7.127
            }
            \def\pointsOutside{
                0.310/5.948, 0.228/6.323, 0.126/6.883, 0.682/5.506, 0.680/6.265, 0.442/4.731, 0.829/6.699, 0.924/7.156, 0.980/5.041, 1.115/6.369, 1.412/5.977, 1.572/5.621, 1.325/5.131, 1.490/7.373, 1.450/6.624
            }
            \def\kInside{
                2.632/5.087, 2.612/7.334, 2.799/5.883, 2.650/6.181, 2.013/6.603, 2.385/5.824, 2.178/5.269,  1.936/5.012, 1.896/7.127
            }
            \def\kOutside{
                0.829/6.699, 0.924/7.156, 0.980/5.041, 1.115/6.369, 1.412/5.977, 1.572/5.621, 1.325/5.131, 1.490/7.373, 1.450/6.624
            }
            
            \foreach \x\y in \pointsOutside{
                \pgfmathsetmacro{\fy}{1.3*sin(deg(\y*3))/(-\y) + 1.75}
                \pgfmathsetmacro{\deltax}{\fy - \x}
                \pgfmathsetmacro{\relativeDistance}{\deltax / \fy};
                \pgfmathsetmacro{\percentageDistance}{\relativeDistance*100}
                \pgfmathsetmacro{\borderPercentage}{\relativeDistance*70}
                \draw[OIlightblue!\borderPercentage!OIblue, fill=OIlightblue!\percentageDistance!OIblue] (\x, \y) circle (0.1);
                \pgfmathsetmacro{\onArrow}{1.75*(1-\relativeDistance)}
                \draw[OIlightblue!\borderPercentage!OIblue, fill=OIlightblue!\percentageDistance!OIblue] (\onArrow, 3.5) circle (0.07);
                    }

            \foreach \x\y in \pointsInside{
                \pgfmathsetmacro{\fy}{1.3*sin(deg(\y*3))/(-\y) + 1.75}
                \pgfmathsetmacro{\deltax}{\x - \fy}
                \pgfmathsetmacro{\relativeDistance}{\deltax / (3.5 - \fy)};
                \pgfmathsetmacro{\percentageDistance}{\relativeDistance*100}
                \pgfmathsetmacro{\borderPercentage}{\relativeDistance*70}
                \draw[OIyellow!\borderPercentage!OIred, fill=OIyellow!\percentageDistance!OIred] (\x, \y) circle (0.1);
                \pgfmathsetmacro{\onArrow}{1.75 + 1.75*\relativeDistance}
                \draw[OIyellow!\borderPercentage!OIred, fill=OIyellow!\percentageDistance!OIred] (\onArrow, 3.5) circle (0.07);
            }
            
            \foreach \xorg\y in \kOutside{
                \pgfmathsetmacro{\x}{\xorg + 4}
                \pgfmathsetmacro{\fy}{1.3*sin(deg(\y*3))/(-\y) + 5.75}
                \pgfmathsetmacro{\deltax}{\fy - \x}
                \pgfmathsetmacro{\relativeDistance}{\deltax / (\fy - 4)};
                \pgfmathsetmacro{\percentageDistance}{\relativeDistance*100}
                \pgfmathsetmacro{\borderPercentage}{\relativeDistance*70}
                \draw[OIlightblue!\borderPercentage!OIblue, fill=OIlightblue!\percentageDistance!OIblue] (\x, \y) circle (0.1);
                \pgfmathsetmacro{\onArrow}{4.0 + 1.75*(1-\relativeDistance)}
                \draw[OIlightblue!\borderPercentage!OIblue, fill=OIlightblue!\percentageDistance!OIblue] (\onArrow, 3.5) circle (0.07);
            }
                    
            \foreach \xorg\y in \kInside{
                \pgfmathsetmacro{\x}{\xorg + 4}
                \pgfmathsetmacro{\fy}{1.3*sin(deg(\y*3))/(-\y) + 5.75}
                \pgfmathsetmacro{\deltax}{\x - \fy}
                \pgfmathsetmacro{\relativeDistance}{\deltax / (7.5 - \fy)};
                \pgfmathsetmacro{\percentageDistance}{\relativeDistance*100}
                \pgfmathsetmacro{\borderPercentage}{\relativeDistance*70}
                \draw[OIyellow!\borderPercentage!OIred, fill=OIyellow!\percentageDistance!OIred] (\x, \y) circle (0.1);
                \pgfmathsetmacro{\onArrow}{5.75 + 1.75*\relativeDistance}
                \draw[OIyellow!\borderPercentage!OIred, fill=OIyellow!\percentageDistance!OIred] (\onArrow, 3.5) circle (0.07);
            }

            \draw [dashed] (4.8, 3.7) rectangle (5.7, 3.3);
            \node [anchor=south west] at (4.8, 3.7) {\footnotesize k=9};
            \draw [dashed] (5.8, 3.7) rectangle (6.8, 3.3);
            \node [anchor=south east] at (6.8, 3.7) {\footnotesize k=9};

        \end{tikzpicture}
    }
    \caption{Schematic of the Improved SortSample algorithm. \textbf{a)} Sample randomly until at least $t$ points inside and $t$ points outside. \textbf{b)} Sort samples by distance to surface. \textbf{c)} Take nearest $k$ samples on each side of the surface.}
    \label{fig:sortsample_schematic}
\end{figure}
    \fi

            We use the \gls{iss}~\cite{henrich2024looc} to generate training data.
            Unlike the original SortSample~\cite{henrich2024registered}, \gls{iss} can produce data for nested objects such as \glspl{roi} inside of a larger object.
            The fundamental idea of \gls{iss} is illustrated in \Cref{fig:sortsample_schematic}.
            Spatial points are uniformly sampled within an extended bounding box of each object segment.
            For each segment, points are sampled until at least $t$ query points are found that are inside the segment and $t$ points are found that are outside the segment.
            These two sets of points are then sorted based on the distance to the nearest surface.
            Only the closest $k$ points to surfaces in each set are kept.
            All such sets from each segment are appended to create a ground truth occupancy point cloud $Q^\text{gt}$.
            The used hyperparameters are listed in \Cref{tab:hyper}.

            \subsection{Simulation based Data Generation:}
                Training samples $(Q^\text{gt},P)$ are created from simulated deformed states of a prior model. 
                Labelled query points $Q^\text{gt}$ are sampled through \gls{iss} and $P$ is generated with a virtual depth camera.
                The position of the virtual camera is randomized to allow reconstructions independent of a specific camera pose. %
                    
                We use the \gls{fem} physics engine \gls{sofa}~\cite{faure2012sofa} for deformation simulation.
                \gls{sofa} is an open-source framework for mechanical simulations, designed with a strong emphasis on biomechanics and robotics to allow for modeling of deformations.
                In addition to our prior experience with \gls{sofa}, it is also used for simulating soft-robotic actuators, which also require precise internal deformation modeling~\cite{Tian2024MultiTap, lai2023sim, makiyeh2023shape}.
                The physical sim-to-real gap of modern \gls{fem} simulators such as \gls{sofa} is narrow enough to enable sim-to-real transfer of deformable object manipulation tasks~\cite{scheikl2023simtorealtransfer}.
                The undeformed, prior 3D surface meshes of the objects of interest are used for generating a sparse hexahedral \gls{fem} topology.
                The objects are then deformed by applying forces on interaction regions on the object surface.
                The deformed states of the objects are then used to create training samples $(Q^\text{gt},P)$.
    \ifrenderfigures

\begin{figure}[ht]
    \centering
    \resizebox{\columnwidth}{!}{
        \begin{tikzpicture}
            \node at (5, 0) {\includegraphics[width=7.5cm]{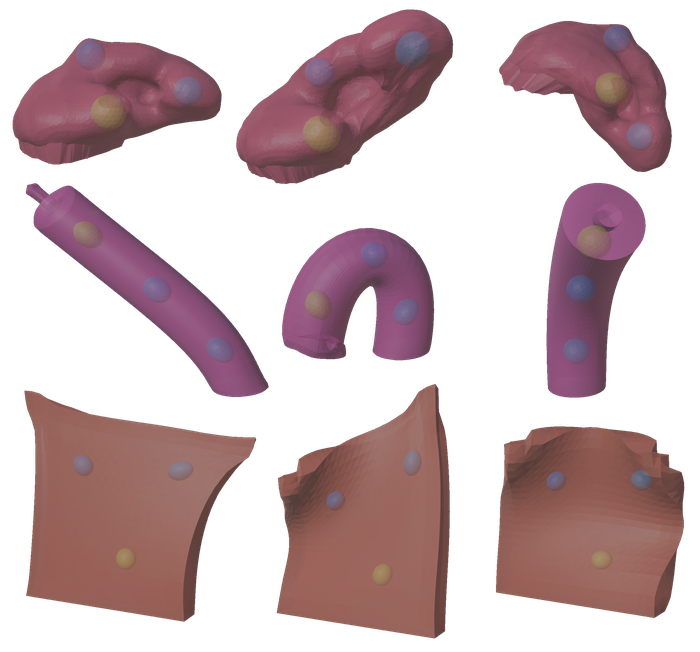}};
            
            \node[rotate=90] at (1, 2.3) {Organ};
            \node[rotate=90] at (1, 0.3) {Cylinder};
            \node[rotate=90] at (1, -2) {Slab};
        \end{tikzpicture}
    }
    \caption{Examples for simulated deformations of Organ, Cylinder, and Slab. The deformations are created by applying forces on random (Organ) or pre-defined regions (Cylinder - Tip, Slab - Corners). The Regions of Interest (ROIs) are represented as deformable spheres inside each object.}
    \label{fig:training_deformation_examples}
\end{figure}
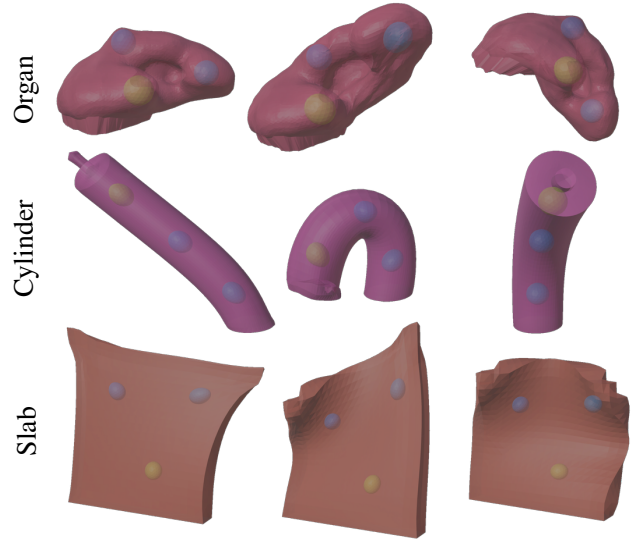

    \fi

                Example deformations for our three evaluation objects \textit{Organ}, \textit{Cylinder}, and \textit{Slab} are shown in \Cref{fig:training_deformation_examples}.
                The \textit{Slab} and \textit{Cylinder} allow us to assess \gls{methodname} in low-feature scenarios, while the \textit{Organ} presents a more complex 3D shape.
                Data generation and training is visualized in \Cref{fig:training_pipeline}, and \Cref{fig:training_progress_visualization} illustrates an example training progress for the \textit{Organ} object.
                The used hyperparameters are listed in \Cref{tab:hyper}.
    \ifrenderfigures
        \input{figures/training_pipeline}
    \fi

    \ifrenderfigures

\definecolor{OIblack}{RGB}{0, 0, 0}
\definecolor{OIgreen}{RGB}{0, 158, 115}
\definecolor{OIblue}{RGB}{0, 114, 178}
\definecolor{OIlightblue}{RGB}{86, 180, 233}
\definecolor{OIyellow}{RGB}{240, 228, 66}
\definecolor{OIorange}{RGB}{230, 159, 0}
\definecolor{OIred}{RGB}{213, 94, 0}
\definecolor{OIpink}{RGB}{204, 121, 167}

\begin{figure}
    \begin{tikzpicture}
        \centering
    
        \begin{axis}[
            width=\columnwidth, height=7cm,
            xlabel={Epoch},
            ylabel={mIoU},
            xmin=0, xmax=720,
            ymin=0.5, ymax=1.0,
            axis lines=left,
            ytick={0.5,0.6,0.7,0.8,0.9,1.0},
        ]

        \draw[thick, OIblack, dashed] (axis cs:170, 0.62) -- (axis cs:1,0.5094) node[fill=OIblue, circle, inner sep=1.5pt] {};
        \node at (axis cs:170, 0.62) {\includegraphics[width=3.2cm, trim=0 200 410 0, clip]{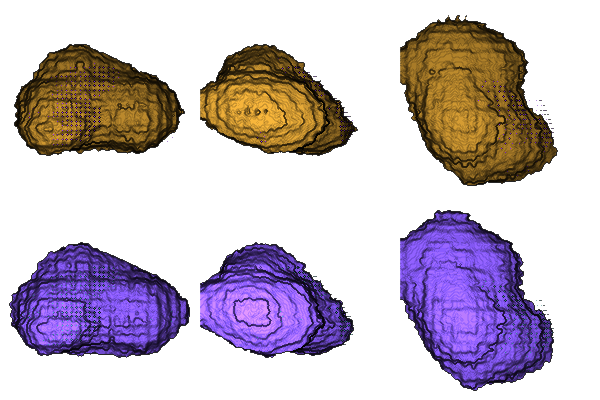}};
        
        \draw[thick, OIblack, dashed] (axis cs:230, 0.82) -- (axis cs:39,0.8149877786636353) node[fill=OIblue, circle, inner sep=1.5pt] {};
        \node at (axis cs:230, 0.82) {\includegraphics[width=3.2cm, trim=0 200 410 0, clip]{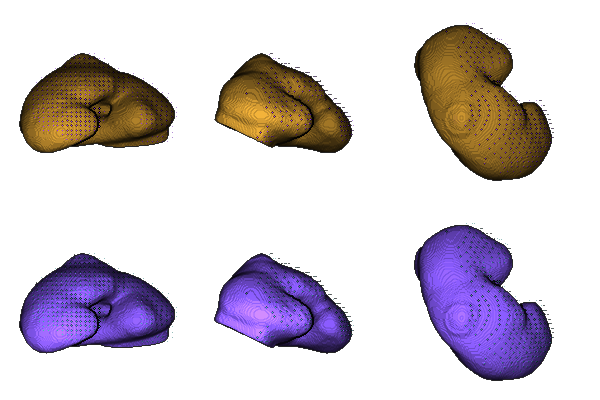}};
    
        \draw[thick, OIblack, dashed] (axis cs:519, 0.7) -- (axis cs:339,0.9587830305099487) node[fill=OIblue, circle, inner sep=1.5pt] {};
        \draw[thick, OIblack, dashed] (axis cs:519, 0.7) -- (axis cs:699,0.9603912234306335) node[fill=OIblue, circle, inner sep=1.5pt] {};
        \node at (axis cs:519, 0.7) {\includegraphics[width=4cm, trim=0 200 410 0, clip]{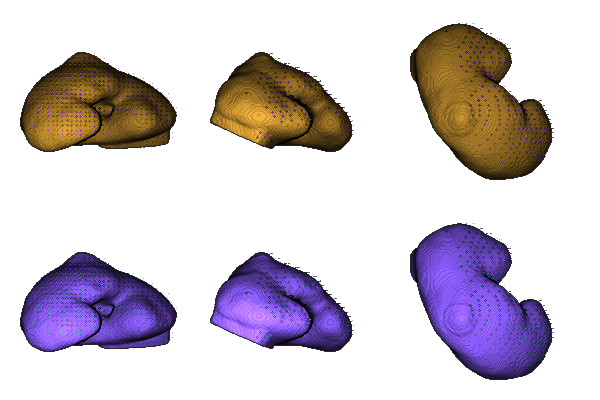}};
        \node at (axis cs:519, 0.94) {almost identical};
        \node at (axis cs:519, 0.98) {Output};

        \addplot[thick, OIblue, smooth] table [x=Step, y=mIoU, col sep=comma] {plots/training_progress/training_progress_export_with_epoch_1.csv};
    
        \end{axis}
    \end{tikzpicture}
        \caption{Training progress for the Organ scene. A dense reconstruction is performed by querying the occupancy network on an equidistant grid.
        We show the reconstruction quality, computed on a separate test dataset, as the mean Intersection over Union (mIoU) over $700$ epochs.
        After approximately $300$ epochs, the network produces outputs that are almost identical to the final outputs. The input point cloud $P$ is included as pink points in the dense reconstruction. One epoch takes approximately one minute to complete on a single Nvidia RTX 4090 (Nvidia, United States).}
    \label{fig:training_progress_visualization}
\end{figure}

    \fi

                \begin{table}[tb]
                    \caption{Hyperparameters for data generation and training.}
                    \label{tab:hyper}
                    \centering
                    \begin{tabular}{l c}
                    \toprule
                    Hyper Parameter & Value \\
                    \midrule
                        Training samples $(Q^\text{gt},P)$ & $30\,000$\\
                        Batch size               & $40$\\
                        Epochs                   & $700$\\
                        SortSample               & $t=k=128$\\
                        Optimizer                & Adam\\
                        Learning rate            & $0.0005$\\
                        Loss function weighting $(\mathcal{L}_{\lambda}$) & $\lambda = 100$\\
                        \bottomrule
                    \end{tabular}
                \end{table}

        \noindent\textbf{Inference:}
            The \gls{on}, conditioned on a point cloud observation, encodes the shape and positions of all objects implicitly. 
            We query the \gls{on} using query points sampled randomly from $[-1.5,1.5]^3$, to obtain the relevant spatial information for downstream tasks.
            The bounding box extends outside the normalization range $[-1,1]^3$ to ensure the full object is reconstructed, see \Cref{sec:methods_reconstruction_using_occupancy_networks}.
            To focus the samples on relevant parts of this space, we apply a two-stage sampling approach.
            First, we query the volume using $10\,000$ query points to obtain a rough bounding box of the objects.
            This bounding box is enlarged by $20\%$ and then sampled more densely to obtain the final dense output point cloud.
            Enlarging the bounding box minimizes the possibility of surfaces being cut-off.
            We experimentally evaluate the number of points required to accurately localize each \gls{roi}.
            A resulting dense 3D point cloud is shown in \Cref{fig:process_overview}.
            This explicit dense output point cloud can then be used for downstream tasks, such as the targeting of specific \glspl{roi} described in \Cref{sec:methods_autonomous_puncturing}.

        \subsection{Local Per-Point Relative Uncertainty Estimation:}
            For safety-critical applications such as surgical interventions, we are interested in how certain the \gls{on} is about its predictions.
            For the application of puncturing an \gls{roi} inside a deformable object, we speculate that targeting the Cartesian position where the \gls{on} is most certain improves the robustness of the puncturing task.
            We therefore extend \glspl{on} to enable the estimation of per-point relative uncertainty.
    
            We consider two methods to compute uncertainty: a naive activation entropy based approach and \gls{mcd}~\cite{pmlr-v48-gal16}.
            
            \paragraph{Activation Entropy}
                The activation entropy of the network's logits, after applying a softmax, can correlate to the model's confidence about its prediction~\cite{kendall2017uncertainties}.
                Lower entropy in the output layer typically corresponds to higher confidence in predictions, as the model is more certain about the specific classes.
                Conversely, higher activation entropy suggests higher uncertainty, indicating that the model does not favor any particular class.
                To this end, we compute the categorical probability distribution
                \begin{equation}
                \textbf{p}_q = \left(\mathcal{M}_\theta(0|q,o), \cdots, \mathcal{M}_\theta(C|q,o)\right),
                \label{eq:prob}
                \end{equation}
                from the softmax outputs for each query point $q$ given some observation $o$ across $C+1$ classes. $C$ is the number of segments and $0$ is reserved for points outside of all segments.
                We then compute the predictive entropy
                \begin{equation}
                            H(\textbf{p}_q) = -\sum_{j} \textbf{p}_q^{(j)} \log(\textbf{p}_q^{(j)} + \epsilon),
                \end{equation}
                where $\textbf{p}_q^{(j)}$ is the probability of point $q$ being of class $j$ and $\epsilon$ is a small value added for numerical stability.
    
            \paragraph{Monte-Carlo Dropout}
                \gls{mcd} is the second approach that we investigate to quantify the uncertainty of neural network predictions~\cite{pmlr-v48-gal16}.
                Dropout is normally used during training to improve the generalization performance of the network, and is usually disabled during inference.
                With \gls{mcd}, dropout is also enabled during inference.
                This results in a virtual ensemble of models that can be used for inference.
                Instead of a single deterministic prediction, we sample the model $30$ times with the same input but different random seeds for dropout.
                We then average the predictions $\textbf{p}_{q,i}$ across all random seeds $i$ to obtain the mean predicted probability for each point as 
                \begin{equation}
                    \hat{\textbf{p}_q} = \frac{1}{30} \sum_{i=1}^{30} \textbf{p}_{q,i}.
                \end{equation}
                Finally, we compute the predictive entropy
                    \begin{equation}
                    H(\hat{\textbf{p}_q}) = -\sum_{j} \hat{\textbf{p}}_q^{(j)} \log(\hat{\textbf{p}}_q^{(j)} + \epsilon)
                    \end{equation}
                using the averaged probabilities $\hat{\textbf{p}}_q$, where $\hat{\textbf{p}}_q^{(j)}$ is the average probability of class $j$, and $\epsilon$ is a small value for numerical stability.

        \subsection{Aggregated Uncertainty Estimation:}
            Per-point relative uncertainty estimation provides information about what parts of a reconstruction the \gls{on} is most confident about.
            However, the per-point uncertainties do not provide global information regarding how certain the overall reconstruction is.
            For safety critical applications, having a single value indicating the confidence or quality of the reconstruction can be important, for example, to prevent a robotic system from performing a task when confidence is lacking or to provide feedback to human operators.
            We investigate whether the activation entropy and \gls{mcd} entropy can also be used to obtain such a single aggregated uncertainty value.
            For this, we aggregate the local per-point uncertainties from a dense output point cloud to obtain a single aggregated uncertainty value. 
            For aggregation, we take the mean of the local per-point uncertainties
            
            \begin{equation}
                H_{\text{aggregated}} = \frac{1}{|Q|} \sum_{q \in Q} H(\textbf{p}_q),
            \end{equation}
            where $ H_{\text{aggregated}} $ is the aggregated uncertainty estimate, $Q$ is the query point cloud, and $ H(\textbf{p}_q) $ is the predictive entropy computed for the query point $q$.
            This provides an estimate of the model's average uncertainty.

        \subsection{Explainability}
            Uncertainty estimation can provide information about how confident the system is about its predictions.
            Explainability clarifies how or why a model makes a specific prediction~\cite{zintgraf2017visualizing,fong2017interpretable,selvaraju2017grad}.
            In medical applications, autonomous systems are supervised by human experts to ensure the safety of the procedure.
            In these settings, explainability may increase transparency, build trust, and simplify the identification and correction of biases and errors.
            We address explainability of our method with a point cloud masking based approach to identify which features in the input point cloud $P$ are most relevant to the prediction, and present the results qualitatively.

        \noindent\textbf{Masking-based Explainability:}
            Given an input point cloud $P$ we want to identify the parts of the point cloud that are most important for the \gls{on} predictions.
            For this task we consider the explainability for reconstructing the whole object with all internal structures.
            To this end, we propose a point cloud masking-based approach.
            Intuitively, the idea is to remove circular patches of the input point cloud and quantify how much the subsequent reconstruction is affected.
            The full point cloud $P$, without any removed patches, serves as the baseline as it contains the maximum amount of information.
            
            We start by sampling $40\,000$ random points in the extended bounding box $U^3(-1.5,1.5)$ to obtain a query point cloud $Q = \{q_1, \cdots, q_{40000}\} \subset \mathbb{R}^3$.
            We then compute the baseline prediction
                \begin{equation}
                    S^{\text{max}} = \mathcal{M}^\text{E}_\theta(Q,P).
                \end{equation}

            \noindent Additionally, we define the neighborhood function to use for circular patch selection
                \begin{equation}
                    N_P(p,r) = \left\{x \in P \mid \Vert p - x\Vert_2 < r\right\},
                \end{equation}
                where $r$ is the radius of the circular patch and $p$ is the point whose neighboring points are to be determined.
                The patches selected by this function will be removed from $P$ to evaluate how important the contained points are for the reconstruction.
           
           \noindent We create the masked point clouds $P_i$ by removing patches for each point in the original point cloud $P$:
                \begin{equation}
                    P_i = P \setminus N_P(p_i,r).
                \end{equation}
            Specifically, for every point in $P$, we generate a masked point cloud by removing the patch around that point.
            This process results in as many masked point clouds as there are points in $P$.
                
            \noindent We then infer dense output point clouds $S_i$ for each masked point cloud $P_i$:
                \begin{equation}
                    S_i = \mathcal{M}^\text{E}_\theta(Q,P_i).
                \end{equation}

            \noindent Finally, we can compute the similarity between $S_i$ and $S^\text{max}$.
            If $S_i$ and $S^\text{max}$ are labelled identically, the missing patch of points is not relevant for reconstruction.
            As a similarity measure $h$, we compute the fraction of points with matching labels.
            We can visualize the importance of a point $p_i$ through a heat-map colorization, where the scalar value of a point $p_i$ is the corresponding similarity value $h(S^\text{max}, S_i)$.
            The explainability values are affected by the radius $r$.
            In preliminary experiments, we found that setting the radius $r$ to $20\%$ of the longest side of the bounding box around the point cloud $P$ yielded good results.

\section{Autonomous Puncturing}
\label{sec:methods_autonomous_puncturing}
    This section outlines the methods to perform autonomous puncturing tasks using the structural information provided by \gls{methodname}.
    The goal is to accurately target and puncture \glspl{roi} within \highly deformable objects based on a single point cloud observation.

    For this, we use a point cloud observation, obtained using our robotic setup, to infer a dense output point cloud.
    The dense output point cloud is transformed into the robot's coordinate system and used to determine the target positions for puncturing and to plan the puncture path.
    Finally, we describe our approach to calibrating the robot to improve its absolute positioning accuracy.

    \subsection{Robotic Setup}
        Our robotic setup consists of a rigid table with a $50\times50$~mm grid of mounting points to which we attach deformable objects.
        Additionally, a Zivid One+ M (Zivid, Norway) depth camera and a UFactory xArm 7 (UFactory, China) industrial robot are attached to the table.
        A needle is attached to the industrial robotic arm, a close up is shown in \Cref{fig:process_overview}.
        The Zivid One+ M is mounted approximately \SI{1}{\m} from the deformable objects.
        The setup is shown in \Cref{fig:setup_first_figure}.

    \subsection{Physical Phantoms}
        \label{sec:physical_phantom_creation}
    \ifrenderfigures
        \input{figures/phantom_creation_T}
    \fi

    \ifrenderfigures

\begin{figure*}[ht]
    \centering
    \begin{minipage}[b]{0.32\textwidth}
        \centering
        \begin{tabular}{@{}c@{}}
            \includegraphics[width=\linewidth]{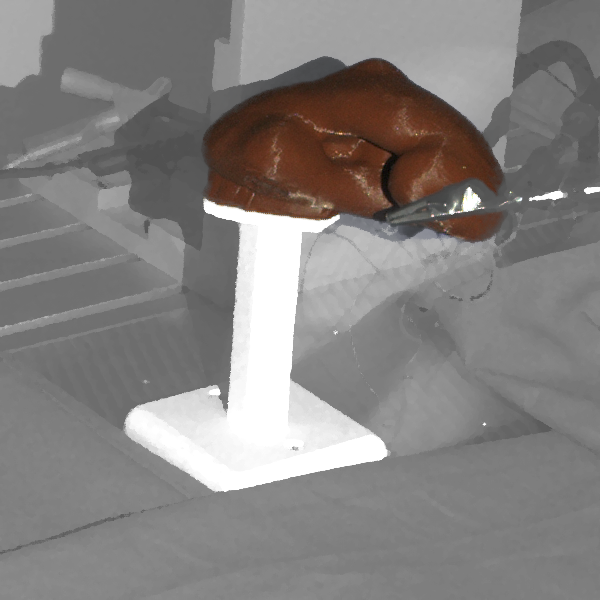} \\[-10pt]
            \includegraphics[width=0.5\linewidth]{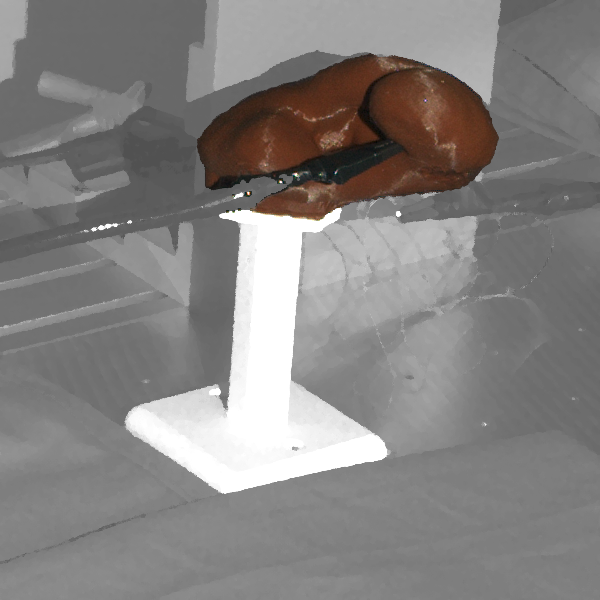}%
            \includegraphics[width=0.5\linewidth]{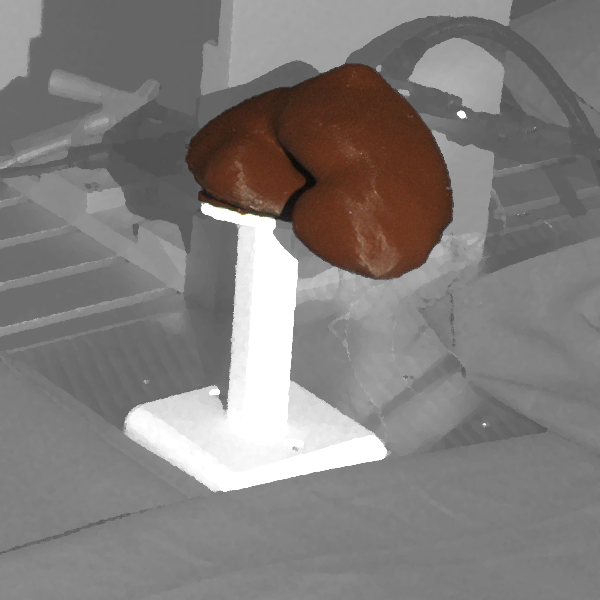} \\
            Organ
        \end{tabular}
    \end{minipage}
    \begin{minipage}[b]{0.32\textwidth}
        \begin{tabular}{@{}c@{}}
            \includegraphics[width=\linewidth]{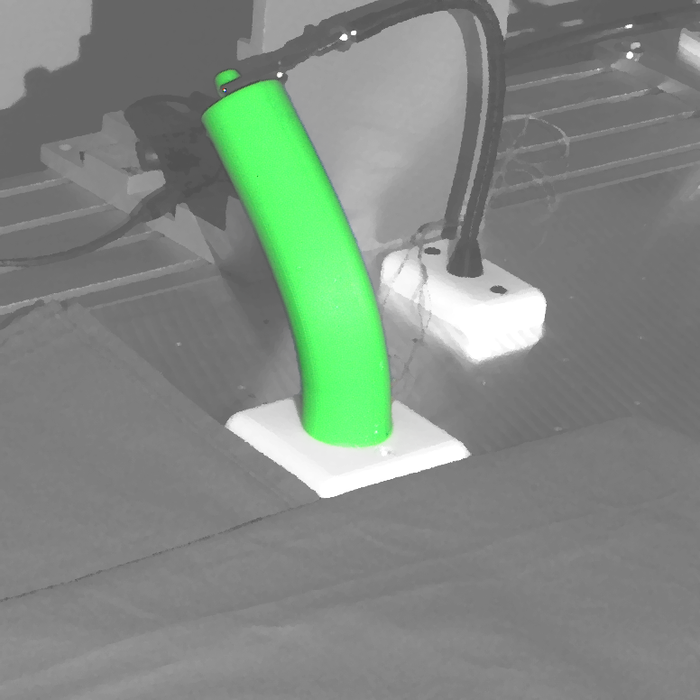} \\[-10pt]
            \includegraphics[width=0.5\linewidth]{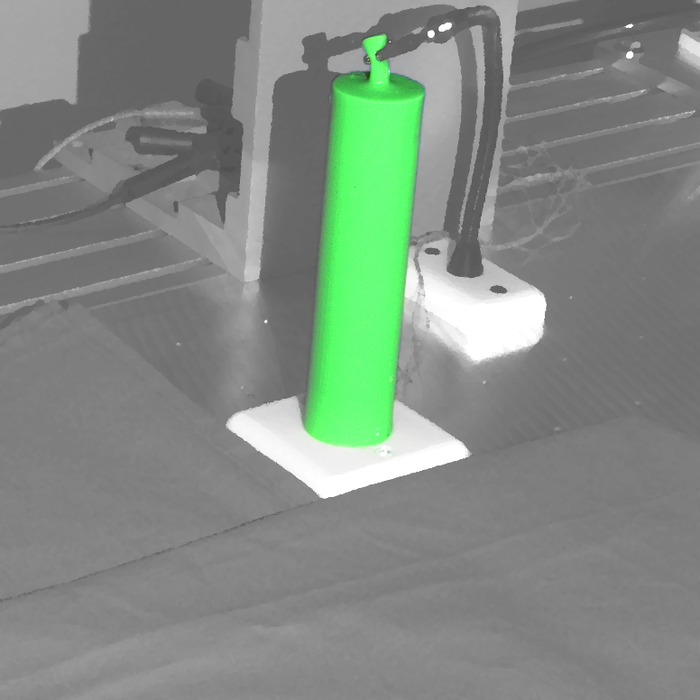}%
            \includegraphics[width=0.5\linewidth]{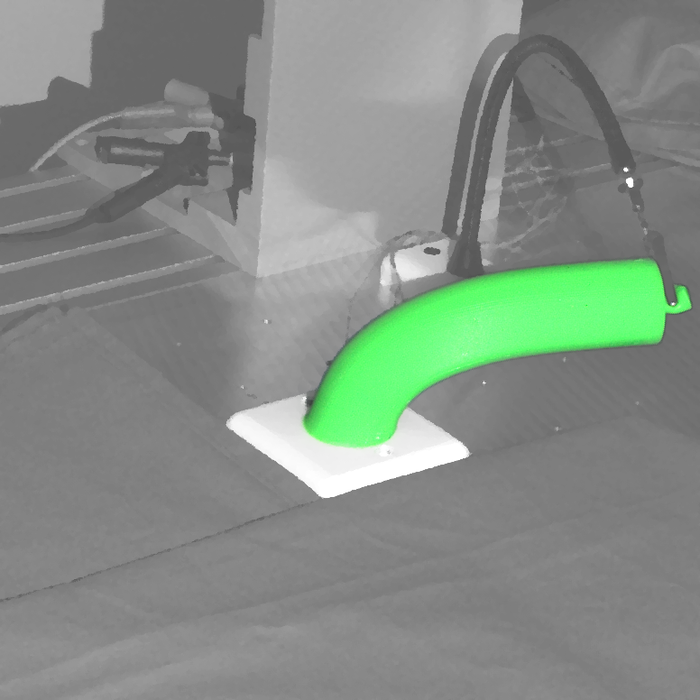} \\
            Cylinder
        \end{tabular}
    \end{minipage}
    \begin{minipage}[b]{0.32\textwidth}
        \begin{tabular}{@{}c@{}}
            \includegraphics[width=\linewidth]{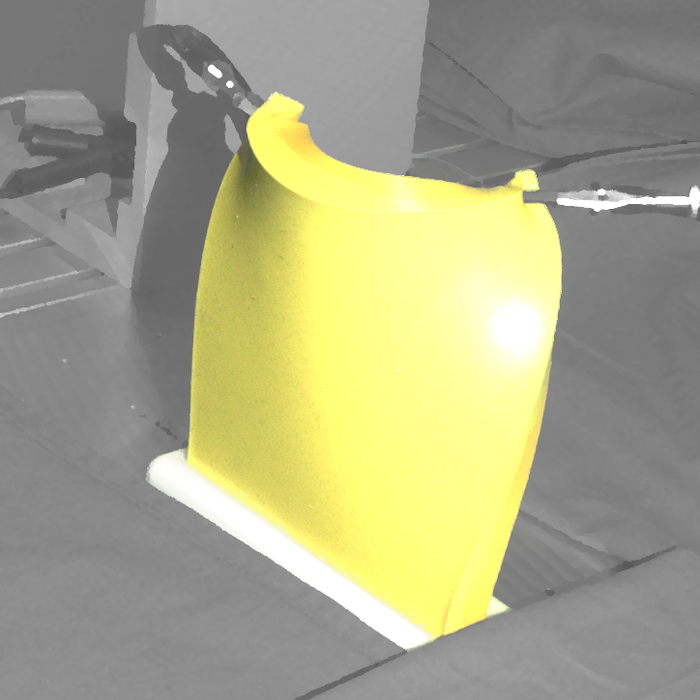} \\[-10pt]
            \includegraphics[width=0.5\linewidth]{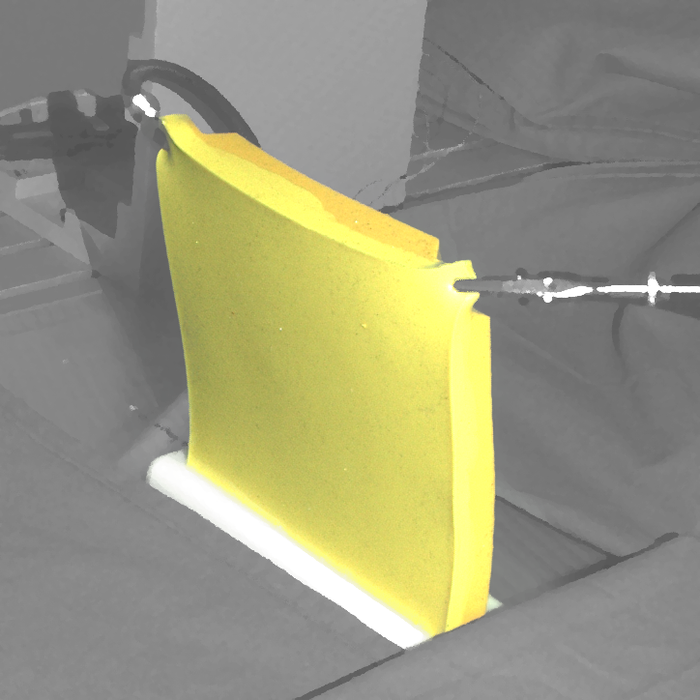}%
            \includegraphics[width=0.5\linewidth]{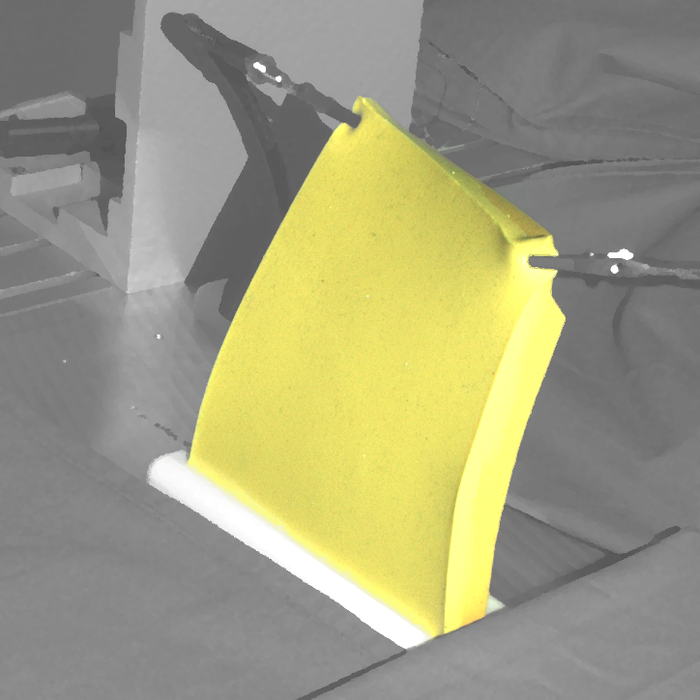} \\
            Slab
        \end{tabular}
    \end{minipage}

    \caption{Example deformations of real-world objects. The images are taken by the depth camera, alongside the actual point clouds used as inference input. Helping hands are used to deform the objects.}
    \label{fig:example_deformations_real_world}
\end{figure*}

    \fi

        We create three different silicone phantoms: \textit{Organ}, \textit{Cylinder}, and \textit{Slab}.
        Each phantom is deformable and contains three spherical \glspl{roi} with a diameter of \SI{17}{\mm} that will be the targets for robotic puncturing.
        Further, the phantoms are re-usable and accurately represent the 3D objects used for training.
        To create our phantoms, we use a silicone molding process with electrically conducting \glspl{roi}, see \Cref{fig:phantom_creation}.
        Whether a puncture is successful can be determined by checking for electrical continuity between the needle and a wire connected to the targeted \gls{roi}.
        We first design and 3D print molds with plugs to define the negative space for the spherical \glspl{roi}.
        We then mix the silicone (ECOFLEX 0020, Smooth-On, United States) with silicone color pigment (Silc Pig, Smooth-On, United States) with a weight ratio of $100:3$.
        The mixture is degassed using a vacuum chamber and poured into the 3D printed mold.
        The plugs that form the negative space for the \glspl{roi} are inserted.
        After the silicone is fully cured, the plugs are removed, leaving negative space for three \glspl{roi}.
        Each spherical \gls{roi} is formed with extra fine steel wool (grade 0000) using another 3D printed mold.
        A thin silicone wire is stripped and entangled with the steel wool.
        The \glspl{roi} are then positioned inside the silicone phantom.
        The wires are threaded through the phantom, such that they exit at the backside of the phantom.
        The negative spaces with the steel wool \glspl{roi} are then filled with silicone.
        The phantom is then placed into the vacuum chamber to ensure the silicone fully infuses each \gls{roi}.
        After topping up the negative spaces as needed and fully curing the silicone, the excess is removed.
        Each phantom is then glued onto a rigid 3D base using a primer and cyanoacrylate adhesives.
        The rigid base allows the phantom to be attached to the table of the experimental setup.
        Examples how the real-world phantoms can be deformed are shown in \Cref{fig:example_deformations_real_world}.

    \subsection{Target Position Prediction}
    \label{sec:target_position_prediction}
        To obtain the structural information of the deformable phantom, we first acquire a point cloud of its surface using our depth camera.
        The phantom surface is segmented from the background by color.
        From the input surface point cloud, the \gls{on} then infers a dense output point cloud that includes both the external shape and internal structures of the deformable phantom.
        We then transform the dense output point cloud into the robot coordinate system for trajectory planning.
        The goal position for puncturing is defined as the center of each \gls{roi}.
        The straightforward way of determining the \gls{roi} centers is calculating the geometric center of points in the point cloud that are labeled as belonging to the \gls{roi}.
        As an alternative, we further investigate the use of local uncertainties and softmax-probabilities to weigh the influence of each point when computing the center.
        
        \gls{uwc} is the centroid of a set of points, using the prediction uncertainty of each point as a weight.
        Given a set of points $\{ (x_i, y_i, z_i, u_i) \}_{i}$ where $(x_i, y_i, z_i)$ are the coordinates of the $i$-th point and $u_i$ is the uncertainty of the predicted label, the \gls{uwc} is a weighted arithmetic mean of the coordinates.
        To compute the \gls{uwc}, we
        \begin{enumerate}
            \item normalize the uncertainties
            \[
            u_i' = \frac{u_i}{\sum_{j} u_j},
            \]
            \item convert uncertainties to certainties
            \[
            c_i = 1 - u_i',
            \]
            \item normalize certainties
            \[
            c_i' = \frac{c_i}{\sum_{j} c_j}, \ \text{and} 
            \]
            \item compute the \gls{uwc}
            \[
            \text{UWC} = \left( \sum_{i} c_i' x_i, \sum_{i} c_i' y_i, \sum_{i} c_i' z_i \right).
            \]
        \end{enumerate}

        Finally, we consider the \gls{spwc}.
        The \gls{spwc} is the centroid of a set of points, using the softmax probabilities of each point as a weight.
        For this we consider the per-point softmax probability of the predicted label as a weight, see \Cref{eq:prob}.
        Similar to the \gls{uwc}, these weights are normalized over all points with the same label.

    \subsection{Puncture Path Planning}
        For puncturing an \gls{roi}, we use the target position as described in \Cref{sec:target_position_prediction}.
        We then locate the point in the dense output point cloud that is closest on the surface of the deformed phantom.
        The vector between the target point and the corresponding closest point on the surface provides the puncture trajectory.
        For robotic execution, we first approach the surface point at a distance, to align the robotic needle with the puncture trajectory, and then translate the needle along the trajectory to reach the target point.
        
    \subsection{Robot Calibration}
        \label{sec:robotic_calibration}
        As the goal of our robotic setup is to puncture \glspl{roi} with a diameter of \SI{17}{\mm}, the accuracy of the robotic system is critical.
        We use an industrial robotic arm with serial kinematics that has a repeatability of $\pm$\SI{0.1}{\mm} (\ie, the ability to return to the same position again and again).
        However the accuracy (\ie, the ability to move to a desired absolute position in space) varies non-linearly in the workspace with average deviations of \SI{5.466}{\mm} and a maximum of \SI{8.752}{\mm}.
        Therefore, these deviations are comparable to the radius of the \glspl{roi} (\SI{8.5}{\mm}).
        To ensure that \glspl{roi} can be hit reliably, the accuracy of the robotic arm needs to be improved.
        For this, we calibrate its \gls{dh} parameters~\cite{denavit1955kinematic} by learning offsets that minimize the difference between the end-effector positions that were determined by the robot's internal forward kinematics and reference positions measured by an optical tracking system, see \Cref{fig:robot_calibration}.

    \ifrenderfigures

\begin{figure}[!ht]
    \centering
    \resizebox{1.0\columnwidth}{!}{
        \begin{tikzpicture}
        
        \node[anchor=south west, inner sep=0] (image) at (0,0) {\includegraphics[width=12cm]{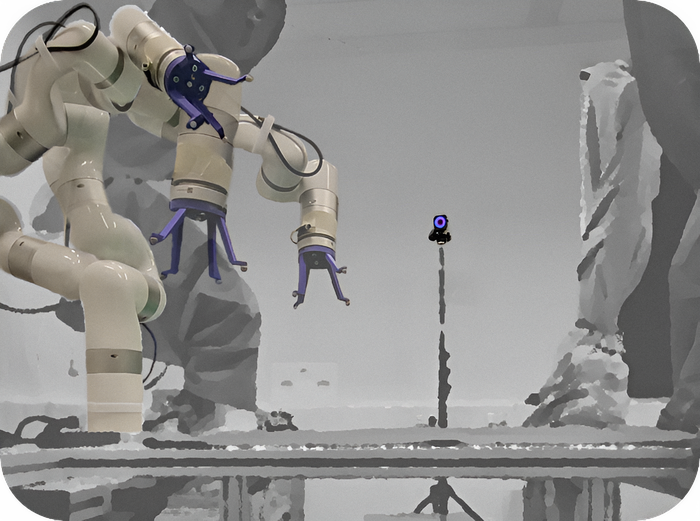}};

        \node[draw, rounded corners=0.5cm, fill=white, text width=2cm, align=center] (PCD)
        at (4.2,1.7) {\includegraphics[width=2cm]{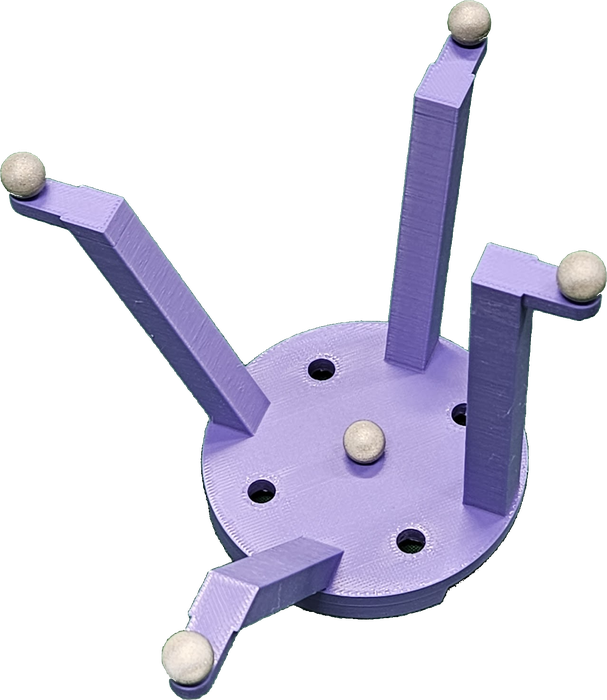}\\\large Marker};

        \draw[white, <-] (3.7,7.7) -- (7.5,7.7) node[pos=1, above] {\large Marker};
        
        \draw[white, <-] (7.55,5.4) -- (7.55,6.1) node[pos=1, above] {\large Tracking Camera};
        
        \node (Zivid) at (10.2,6.5) [circle,scale=1.0] {};

        \end{tikzpicture}
    }
    \caption{
        Calibration setup used to improve the accuracy of the robotic system.
        The xArm7 robot (UFactory, China) with an infrared \textit{Marker} attachment moves around while being tracked by three OptiTrack PrimeX 13W \textit{Tracking Cameras} (NaturalPoint, United States).
        The \textit{Tracking Cameras} are arranged in a triangular configuration around the robotic setup at a distance of approximately \SI{2}{\m}.
        Only one of the \textit{Tracking Cameras} is visible in the image.}
    \label{fig:robot_calibration}
\end{figure}

    \fi

        We describe the kinematic chain of the $7$ \gls{dof} serial robotic arm with one tuple of \gls{dh} parameters $\omega_i := (\theta_i, d_i, \alpha_i, a_i)$ per joint $i$.
        Each tuple of \gls{dh} parameters $\omega_i = \hat{\omega}_i + \Delta\omega_i$ is the sum of manufacturer provided \gls{dh} parameters $\hat{\omega}_i$ and a calibration offset $\Delta\omega_i$ that we estimate from data.
        We further use a position vector $p_{\text{base}} \in \mathbb{R}^3$ and XYZ Euler angles $\phi_{\text{base}} \in \mathbb{R}^3$ to describe the pose of the robot's base in relation to a fixed world-reference coordinate system. 
        We optimize $p_{\text{base}}$, $\phi_{\text{base}}$, and $\Delta\omega_i$ to minimize the error between the robot's position determined by the manufacturer's \gls{dh} parameters $\omega_i$ and its true pose in 3D space.
        To obtain the true position and rotation of the end-effector, we use an infrared marker based tracking system (OptiTrack PrimeX 13W, NaturalPoint, United States) with a 3D printed marker attachment for the robot.
        In our setup, the tracking system achieves an accuracy of approximately \SI{0.4}{\mm}.
        Our calibration dataset consists of triplets $(p_{\text{tracked}}, q_{\text{tracked}}, \xi)$, where $p_{\text{tracked}} \in \mathbb{R}^3$ and $q_{\text{tracked}} \in \mathbb{R}^4$ represent the end-effector position and orientation, respectively, as measured by the optical tracker, while $\xi \in \mathbb{R}^7$ contains the joint values recorded from the robot's encoders.
        The robot moves along trajectories between $500$ randomly generated end-points.
        The triplets are recorded at static end-points of trajectories after a short pause to minimize settling vibrations.

        The complete forward kinematics transformation that describes the pose of the end-effector in relation to the fixed world-reference coordinate system is 
        \begin{equation}
            T_{\text{EE}} = T_{\text{base}} \prod_{i=1}^{7} T_i,
        \end{equation}
        where $T_i \in \mathbb{R}^{4\times4}$ is the homogeneous transformation matrix of the $i$-th joint for \gls{dh} parameters $\omega_i$, constructed as
        \begin{equation}
            T_i = \begin{bmatrix}
            c\theta_i & -s\theta_i & 0 & a_i \\
            s\theta_i\cdot c\alpha_i & c\theta_i \cdot c\alpha_i & -s\alpha_i & -d_i\cdot s\alpha_i\\
            s\theta_i\cdot s\alpha_i & c\theta_i \cdot s\alpha_i & c\alpha_i & d_i\cdot c\alpha_i\\
            0 & 0 & 0 & 1
            \end{bmatrix},
        \end{equation}
        with shorthands $s$ and $c$ for $\sin(\cdot)$ and $\cos(\cdot)$, respectively.
        $T_{\text{base}} \in \mathbb{R}^{4\times4}$ is the homogeneous transformation matrix for $p_{\text{base}}$ and $\phi_{\text{base}}$.
        For optimization, we compute the end-effector position using the forward kinematics with $\theta = \xi + \Delta\theta$, where $\xi$ are the joint values from our dataset.
        The resulting homogeneous matrix $T_{EE}$ encodes the end-effector position $p_{\text{forward}}$ and end-effector orientation quaternion $q_{\text{forward}}$.
        The parameters $p_{\text{base}}$, $\phi_{\text{base}}$, and $\Delta\omega_i$ are optimized by minimizing a combined loss function
        \begin{equation}
            \mathcal{L} = \mathcal{L}_{\text{pos}} + \lambda \mathcal{L}_{\text{rot}},
        \end{equation}
        where $\lambda$ is a weighting for the rotation loss.
        We set $\lambda = 100$ to scale both loss components to the same order of magnitude.
              
        The position loss 
        \begin{equation}
            \mathcal{L}_{\text{pos}} = \frac{1}{S}\sum_{j=1}^{S} \left\| p_{\text{tracked}}^{(j)} - p_{\text{forward}}^{(j)} \right\|_2
        \end{equation}
        is the mean 2-norm between measured end-effector positions $p_{\text{tracked}}$ from the optical tracking system and the predicted position $p_{\text{forward}}$ over $S=500$ samples in the dataset.
        
        The rotation loss

        \begin{equation}
            \mathcal{L}_{\text{rot}} = \frac{1}{S}\sum_{j=1}^{S} 
            \min
            \left\{
            \left\| 
            q_{\text{tracked}}^{(j)} \pm q_{\text{forward}}^{(j)}
            \right\|_2
            \right\}
        \end{equation}
        
        is the mean over the quaternion difference defined by Huynh~\cite{huynh2009metrics}.
        Quaternions can represent the same rotation with either a positive or negative sign.
        The loss function takes the minimum 2-norm difference between both possibilities to account for this sign ambiguity in quaternion representation.
        The optimization is performed with stochastic mini-batch gradient descent, using PyTorch's implementation of the Adam optimizer~\cite{kingma2015adam}.

\section{Experiments}
\label{sec:experiments}
    To evaluate the performance of \gls{methodname} in the context of low-latency targeting of internal structures within \highly deformable objects, we conduct a series of experiments designed to assess various aspects of the system.
    This includes inference speed, prediction accuracy, uncertainty estimation, explainability, robot calibration, a baseline comparison, and real-world autonomous puncturing.
    The centroid error metric measures the error between the predicted and ground truth centroid of an \gls{roi}.
    All time measurements were performed using a single Nvidia RTX 4090 (Nvidia, United States).
    \gls{methodname} was trained according to \Cref{sec:methods_reconstruction} using the prior 3D surface models, containing the \glspl{roi}, of each phantom.
    The hyperparameters were chosen according to \Cref{tab:hyper} unless stated otherwise.

    \subsection{Inference Time}
    We measure \gls{methodname}’s inference speed by varying the number of query points and observing the trade-off between response time and accuracy in localizing internal structures.
    The goal is to optimize speed without sacrificing precision.

    \subsection{Uncertainty Estimation}
    We explore activation-based and \gls{mcd}-based uncertainty estimation.
    These experiments are designed to determine whether per-point uncertainty can be used to influence the centroid estimation of the \glspl{roi} to improve targeting accuracy.
    For this, we compare the centroid error of the \gls{uwc} method with a regular geometric centroid and the \gls{spwc}.
    Additionally, we want to evaluate whether it is possible to obtain a single aggregated uncertainty value.
    This single value can be used to determine if it is safe to perform a task given the input data.
    These experiments are performed in simulation, using $N=50$ different deformations for each object.

    \subsection{Explainability}
    We test the system’s explainability to highlight important regions in the input point cloud.
    This helps identify which features are most relevant for understanding the objects' deformations and ensures that the system’s decisions are interpretable.

    \subsection{Robot Calibration}
    \label{sec:experiments_robot_calibration}
    We evaluate the end-effector positioning error, defined as the distance between the end-effector position predicted by the robot's forward kinematics and the actual position measured by the optical tracker.
    In addition to the calibration as described in \Cref{sec:robotic_calibration}, we perform ablative experiments to show the impact of the individual design decisions.
    The ablations are:
    \begin{itemize}
        \item \textbf{($\Delta\theta$-Only Calibration)} Learning only $\Delta\theta$, to determine whether the error stems from incorrectly calibrated zero positions of the rotational joints in the kinematic chain.
        \item \textbf{(Fixed $T_{\text{base}}$)} Using a manually measured fixed transformation between robot base and world-reference $T_{\text{base}}$.
        \item \textbf{(Dynamic Data)} Collecting data triplets $(p_{\text{tracked}}, q_{\text{tracked}}, \xi)$ during the movement between end-points (\ie, not just static end points).
        \item \textbf{(Single Joint Movement)} Instead of moving along trajectories between end-points in Cartesian space, move the robot in joint space, a single joint at a time.
    \end{itemize}

    \subsection{Comparison to Baseline}
    We compare \gls{methodname} to \gls{v2s}~\cite{pfeiffer2020non}.
    This comparison requires multiple considerations due to the limitations of current deformable registration approaches.
    Current deformable registration methods commonly require an initial rigid alignment of the observation to the prior model~\cite{deng2022survey}.
    \gls{v2s} is able to perform deformable registration without an initial rigid alignment if the object that is to be registered is fully observable (not just a single view).
    In cases where only a partial observation of the surface is present, the results depend highly on the initial rigid alignment.
    Rigid initial alignment approaches, for example using \gls{icp}, will inevitably fail once deformations become too strong.
    Therefore, given single-view or partial observations, the \gls{v2s} authors propose performing the initial alignment manually.
    For this evaluation, we ensure that the evaluation dataset is already rigidly aligned for \gls{v2s}.
    We deform the objects but do not apply any additional rotations or translations.
    Since the same assumption underlies the training data generation (\ie, \gls{v2s} learns to deform already aligned shapes), the alignment distribution is the same as that of the training set.
    This ensures that the observations are optimally aligned with the prior model for \gls{v2s}.
    Finally, we evaluate \gls{v2s} using both single-view and full surface observation of the deformed object.
    Having a full surface observation of the deformable objects is impractical in most applications.
    We will only provide \gls{methodname} with single-view observations and no initial alignment.
    \gls{methodname} is trained from scratch for each object.
    For inference, \gls{methodname} uses $40\,000$ query points.
    Similar to \gls{methodname}, \gls{v2s} is a data-driven method.
    Therefore, we evaluate the benefit of fine-tuning the author provided pretrained \gls{v2s} on our data.
    For evaluation, we train \gls{methodname} and \gls{v2s} on $20\,000$ deformations of each object for $100$ epochs.
    To compare the approaches, we again use the centroid error.
    The test dataset consists of $64$ deformations for each object.
    Additionally, we discuss the training time, inference time, and memory requirements.
    
    \subsection{Real-World Robotic Puncturing}
    In the final set of experiments, we test the overall performance of \gls{methodname} in a real-world robotic puncturing scenario.
    Using deformable silicone phantoms with embedded \glspl{roi}, we task the robot to autonomously puncture the \gls{roi} centers.
    Each phantom is deformed $N=10$ times, examples are shown in \Cref{fig:example_deformations_real_world}.
    We apply the deformations by pushing or pulling the phantoms using helping hands.
    Additionally, we ensure that all deformations are visually distinguishable from each other.
    After each deformation, each of the three \glspl{roi} is targeted for puncture.
    These experiments simulate scenarios such as robotic biopsies and evaluate the practicality of \gls{methodname} on \highly deformed objects for providing structural information to precisely target internal structures.

\section{RESULTS}
\label{sec:results}
    \subsection{Inference Time}
    \ifrenderfigures

\definecolor{OIblack}{RGB}{0, 0, 0}
\definecolor{OIgreen}{RGB}{0, 158, 115}
\definecolor{OIblue}{RGB}{0, 114, 178}
\definecolor{OIlightblue}{RGB}{86, 180, 233}
\definecolor{OIyellow}{RGB}{240, 228, 66}
\definecolor{OIorange}{RGB}{230, 159, 0}
\definecolor{OIred}{RGB}{213, 94, 0}
\definecolor{OIpink}{RGB}{204, 121, 167}

\colorlet{Cent}{OIblue}
\colorlet{Act}{OIgreen}
\colorlet{MC}{OIorange}
\colorlet{Grad}{OIpink}

\def\myplotwidth{0.42\columnwidth}
\def\myplotheight{0.45\columnwidth}

\def\myrceymins{0.0}
\def\myrceymaxs{3.0}

\def\myitymin{0.0}
\def\myitymax{0.1}

\begin{figure}
    \centering

    \begin{tikzpicture}
        \begin{axis}[
            name=Organ,
            width=\myplotwidth,
            height=\myplotheight,
            ymin=\myrceymins,
            ymax=\myrceymaxs,
            legend style={
                at={(1.5,1.03)},
                font=\small,
                anchor=south,
                legend columns=4
            },
            ylabel={ROI Center Error [mm]},
            ylabel style={
                at={(-0.22, 0.5)},
                font=\small,
            },
            ymajorgrids=true,
            yminorgrids=true,
            major grid style={semithick, black!30!white, dashed},
            minor grid style={ultra thin, black!20!white, dashed},
            minor y tick style = transparent,
            x tick label style={
                font=\footnotesize,
                /pgf/number format/fixed,
                /pgf/number format/precision=5
            },
            scaled x ticks=false,
            xtick={0, 50000, 100000, 150000},
            xticklabels={0, 50k, 100k, 150k},
            y tick label style={
                font=\footnotesize,
            },
        ]
            \addplot[
                color=Cent,
                mark=star,
                thick,
            ]
            table [
                x={Number of Query Points},
                y expr={10 * \thisrow{Average Geometric Prediction Error Mean (Last 3 Classes)}},
                col sep=comma
            ] {plots/uncertainty/query_points_vs_inference_time_and_accuracy/organ_biopsy_zivid.csv};
            \addlegendentry{ROI Center Error}

            \addlegendimage{color=Act, mark=square, thick}
            \addlegendentry{Inference Time}
        \end{axis}

        \begin{axis}[
            width=\myplotwidth,
            height=\myplotheight,
            ymin=\myitymin,
            ymax=\myitymax,
            xticklabel=\empty,
            xtick=\empty,
            ytick={0, 0.1, 0.2},
            y tick label style={
                font=\footnotesize,
            },
            yticklabels={,,},
        ]
            \addplot[
                color=Act,
                mark=square,
                thick,
            ]
            table [
                x={Number of Query Points},
                y={Average PCD Generation Time (s)},
                col sep=comma
            ] {plots/uncertainty/query_points_vs_inference_time_and_accuracy/organ_biopsy_zivid.csv};
        \end{axis}

        \begin{axis}[
            name=Cylinder,
            width=\myplotwidth,
            height=\myplotheight,
            at=(Organ.south east),
            anchor=left of south west,
            ymin=\myrceymins,
            ymax=\myrceymaxs,
            xlabel style={
                yshift=3,
                font=\small,
                anchor=north,
            },
            xlabel={Query Points},
            ymajorgrids=true,
            yminorgrids=true,
            major grid style={semithick, black!30!white, dashed},
            minor grid style={ultra thin, black!20!white, dashed},
            minor y tick style = transparent,
            x tick label style={
                font=\footnotesize,
                /pgf/number format/fixed,
                /pgf/number format/precision=5
            },
            scaled x ticks=false,
            xtick={0, 50000, 100000, 150000},
            xticklabels={0, 50k, 100k, 150k},
            y tick label style={
                font=\footnotesize,
            },
            yticklabels=\empty
        ]
            \addplot[
                color=Cent,
                mark=star,
                thick,
            ]
            table [
                x={Number of Query Points},
                y expr={10 * \thisrow{Average Geometric Prediction Error Mean (Last 3 Classes)}},
                col sep=comma
            ] {plots/uncertainty/query_points_vs_inference_time_and_accuracy/cylinder_v2_zivid.csv};
        \end{axis}

        \begin{axis}[
            width=\myplotwidth,
            height=\myplotheight,
            at=(Cylinder.south west),
            anchor=left of south west,
            ymin=\myitymin,
            ymax=\myitymax,
            xlabel style={
                at={(1.65, -0.15)},
                font=\small,
                anchor=north,
            },
            ylabel=\empty,
            axis y line*=right,
            xticklabel=\empty,
            xtick=\empty,
            ytick={0, 0.1, 0.2},
            y tick label style={
                font=\footnotesize,
            },
            yticklabels=\empty
        ]
            \addplot[
                color=Act,
                mark=square,
                thick,
            ]
            table [
                x={Number of Query Points},
                y={Average PCD Generation Time (s)},
                col sep=comma
            ] {plots/uncertainty/query_points_vs_inference_time_and_accuracy/cylinder_v2_zivid.csv};
        \end{axis}

        \begin{axis}[
            name=Slab,
            width=\myplotwidth,
            height=\myplotheight,
            at=(Cylinder.south east),
            anchor=left of south west,
            ymin=\myrceymins,
            ymax=\myrceymaxs,
            ymajorgrids=true,
            yminorgrids=true,
            major grid style={semithick, black!30!white, dashed},
            minor grid style={ultra thin, black!20!white, dashed},
            minor y tick style = transparent,
            x tick label style={
                font=\footnotesize,
                /pgf/number format/fixed,
                /pgf/number format/precision=5
            },
            scaled x ticks=false,
            xtick={0, 50000, 100000, 150000},
            xticklabels={0, 50k, 100k, 150k},
            y tick label style={
                font=\footnotesize,
            },
            yticklabels={,,},
        ]
            \addplot[
                color=Cent,
                mark=star,
                thick,
            ]
            table [
                x={Number of Query Points},
                y expr={10 * \thisrow{Average Geometric Prediction Error Mean (Last 3 Classes)}},
                col sep=comma
            ] {plots/uncertainty/query_points_vs_inference_time_and_accuracy/slab_zivid.csv};
        \end{axis}

        \begin{axis}[
            width=\myplotwidth,
            height=\myplotheight,
            at=(Slab.south west),
            anchor=left of south west,
            ymin=\myitymin,
            ymax=\myitymax,
            xlabel style={
                at={(1.65, -0.15)},
                font=\small,
                anchor=north,
            },
            ylabel={Inference Time [s]},
            ylabel style={
                at={(1.22, 0.5)},
                font=\small,
                anchor=north,
            },
            axis y line*=right,
            xticklabel=\empty,
            xtick=\empty,
            ytick={0, 0.1, 0.2},
            y tick label style={
                font=\footnotesize,
            },
        ]
            \addplot[
                color=Act,
                mark=square,
                thick,
            ]
            table [
                x={Number of Query Points},
                y={Average PCD Generation Time (s)},
                col sep=comma
            ] {plots/uncertainty/query_points_vs_inference_time_and_accuracy/slab_zivid.csv};
        \end{axis}

        \node[anchor=north, outer sep=0, inner sep=0] at ($(Organ.south) + (0, -0.85cm)$) {\small a) Organ};
        \node[anchor=north, outer sep=0, inner sep=0] at ($(Cylinder.south) + (0, -0.85cm)$) {\small b) Cylinder};
        \node[anchor=north, outer sep=0, inner sep=0] at ($(Slab.south) + (0, -0.85cm)$) {\small c) Slab};
        
    \end{tikzpicture}
    \caption{Relationship between the number of query points, the inference time (excludes depth image capture time) and the error of the estimated centroid to the ground truth centroid. Measurements taken on a single Nvidia RTX 4090 (Nvidia, United States).}
    \label{plot:time_vs_accuracy}
\end{figure}
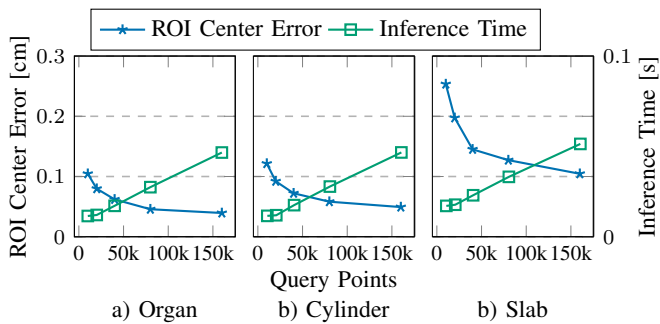

    \fi

    The number of query points used for inference linearly increases inference time, but reduces the centroid error, see \Cref{plot:time_vs_accuracy}.
    Using $20\,000$ query points achieves centroid errors below \SI{2}{\mm} across all objects, with inference times under \SI{20}{\ms}. 
    Increasing the number of points beyond $80\,000$ yields only minimal accuracy gains.
    The most pronounced accuracy improvement relative to additional inference time occurs when increasing the point count from $20\,000$ to $40\,000$, maintaining an inference time of less than \SI{30}{\ms}.
    In comparison to the other objects, \textit{Slab} shows slower inference times.
    This is caused by the larger number of points in the input point clouds, resulting in a slower latent distillation.
    An average input point cloud for \textit{Organ} and \textit{Cylinder} contains $500$ points, whereas \textit{Slab} can contain over $2000$ points, due to the large flat surface the object.

    \subsection{Uncertainty Estimation}
        
        We evaluate the effect of \gls{mcd} on the model performance by comparing it to models trained without dropout.
        A dropout of $20\%$ in the \gls{mlp} part of our architecture has a minimal effect on the centroid error for all objects (less than \SI{0.5}{\mm}).

    \ifrenderfigures

\definecolor{colorcylinder}{rgb}{0.1,0.5,0.1}
\definecolor{colororgan}{rgb}{0.5,0.1,0.1}
\definecolor{colorslab}{rgb}{0.5,0.5,0.1}

\definecolor{OIblack}{RGB}{0, 0, 0}
\definecolor{OIgreen}{RGB}{0, 158, 115}
\definecolor{OIblue}{RGB}{0, 114, 178}
\definecolor{OIlightblue}{RGB}{86, 180, 233}
\definecolor{OIyellow}{RGB}{240, 228, 66}
\definecolor{OIorange}{RGB}{230, 159, 0}
\definecolor{OIred}{RGB}{213, 94, 0}
\definecolor{OIpink}{RGB}{204, 121, 167}

\colorlet{Cent}{OIblue}
\colorlet{Act}{OIgreen}
\colorlet{MC}{OIorange}
\colorlet{Grad}{OIpink}

\def\myplotwidth{0.45\columnwidth}
\def\myplotheight{0.45\columnwidth}

\definecolor{organred}{HTML}{FFF0F0}
\definecolor{cylindergreen}{HTML}{F0FFF0}
\definecolor{slabyellow}{HTML}{FFFFF0}

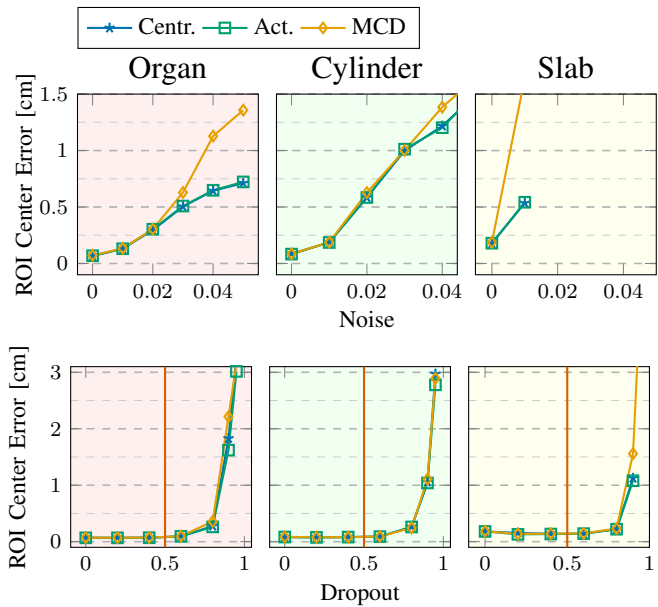
\begin{figure}
  \centering
  \vspace{1.5mm}
  \begin{tikzpicture}
    \begin{axis}[
      name=Organ,
      width=\myplotwidth,
      height=\myplotheight,
      ymin=-1,         %
      ymax=15,         %
      ylabel={ROI Center Error [mm]},
      ylabel style={
        at={(-0.18, 0.5)},
        font=\small,
      },
      xlabel={Noise},
      xlabel style={
        at={(1.60, -0.15)},
        font=\small,
        anchor=north,
      },
      minor ytick={0,2.5,...,15},        %
      ymajorgrids=true,
      yminorgrids=true,
      major grid style={semithick, black!30!white, dashed},
      minor grid style={ultra thin, black!20!white, dashed},
      minor y tick style = transparent,
      x tick label style={
        font=\footnotesize,
        /pgf/number format/fixed,
        /pgf/number format/precision=5,
      },
      y tick label style={
        font=\footnotesize,
      },
      scaled x ticks=false,
      axis background/.style={fill=organred},
      legend style={
        at={(0.0,1.25)},
        font=\small,
        anchor=south west,
        legend columns=4
      },
    ]

      \addplot[
        color=Cent, 
        mark=star, 
        thick
      ] table[
        x index=0,
        y expr=\thisrowno{1}*10,
        y error expr=\thisrowno{2}*10,
        col sep=comma
      ] {plots/uncertainty_without_qnorm/test_noise/raw_data/organ_biopsy_zivid_pcd.csv};
      \addlegendentry{GC}

      \addplot[
        color=Act, 
        mark=square, 
        thick
      ] table[
        x index=0,
        y expr=\thisrowno{1}*10,
        y error expr=\thisrowno{2}*10,
        col sep=comma
      ] {plots/uncertainty_without_qnorm/test_noise/raw_data/organ_biopsy_zivid_pcd_uncertainty_activation.csv};
      \addlegendentry{Act.‐UWC}

      \addplot[
        color=MC, 
        mark=diamond, 
        thick
      ] table[
        x index=0,
        y expr=\thisrowno{1}*10,
        y error expr=\thisrowno{2}*10,
        col sep=comma
      ] {plots/uncertainty_without_qnorm/test_noise/raw_data/organ_biopsy_zivid_pcd_uncertainty_mcdropout.csv};
      \addlegendentry{MCD‐UWC}

      \addplot[
        color=Grad, 
        mark=triangle, 
        thick
      ] table[
        x index=0,
        y expr=\thisrowno{1}*10,
        y error expr=\thisrowno{2}*10,
        col sep=comma
      ] {plots/uncertainty_without_qnorm/test_noise/raw_data/organ_biopsy_zivid_pcd_activation_probability.csv};
      \addlegendentry{SPWC}

    \end{axis}

    \begin{axis}[
      name=Cylinder,
      at=(Organ.south east),
      anchor=left of south west,
      width=\myplotwidth,
      height=\myplotheight,
      ymin=-1,         %
      ymax=15,         %
      ylabel style={
        at={(-0.08, 0.5)},
      },
      yticklabels={,,},
      minor ytick={0,2.5,...,15},
      ymajorgrids=true,
      yminorgrids=true,
      major grid style={semithick, black!30!white, dashed},
      minor grid style={ultra thin, black!20!white, dashed},
      minor y tick style = transparent,
      x tick label style={
        font=\footnotesize,
        /pgf/number format/fixed,
        /pgf/number format/precision=5
      },
      scaled x ticks=false,
      axis background/.style={fill=cylindergreen},
    ]

      \addplot[
        color=Cent, 
        mark=star, 
        thick
      ] table[
        x index=0,
        y expr=\thisrowno{1}*10,
        y error expr=\thisrowno{2}*10,
        col sep=comma
      ] {plots/uncertainty_without_qnorm/test_noise/raw_data/cylinder_v2_zivid_pcd.csv};

      \addplot[
        color=Act, 
        mark=square, 
        thick
      ] table[
        x index=0,
        y expr=\thisrowno{1}*10,
        y error expr=\thisrowno{2}*10,
        col sep=comma
      ] {plots/uncertainty_without_qnorm/test_noise/raw_data/cylinder_v2_zivid_pcd_uncertainty_activation.csv};

      \addplot[
        color=MC, 
        mark=diamond, 
        thick
      ] table[
        x index=0,
        y expr=\thisrowno{1}*10,
        y error expr=\thisrowno{2}*10,
        col sep=comma
      ] {plots/uncertainty_without_qnorm/test_noise/raw_data/cylinder_v2_zivid_pcd_uncertainty_mcdropout.csv};

      \addplot[
        color=Grad, 
        mark=triangle, 
        thick
      ] table[
        x index=0,
        y expr=\thisrowno{1}*10,
        y error expr=\thisrowno{2}*10,
        col sep=comma
      ] {plots/uncertainty_without_qnorm/test_noise/raw_data/cylinder_v2_zivid_pcd_activation_probability.csv};

    \end{axis}

    \begin{axis}[
      name=Slab,
      at=(Cylinder.south east),
      anchor=left of south west,
      width=\myplotwidth,
      height=\myplotheight,
      ymin=-1,         %
      ymax=15,         %
      xmax=0.05,
      xtick distance=0.02,
      yticklabels={,,},
      minor ytick={0,2.5,...,15},
      ymajorgrids=true,
      yminorgrids=true,
      major grid style={semithick, black!30!white, dashed},
      minor grid style={ultra thin, black!20!white, dashed},
      minor y tick style = transparent,
      x tick label style={
        font=\footnotesize,
        /pgf/number format/fixed,
        /pgf/number format/precision=5
      },
      scaled x ticks=false,
      axis background/.style={fill=slabyellow},
    ]

      \addplot[
        color=Cent, 
        mark=star, 
        thick
      ] table[
        x index=0,
        y expr=\thisrowno{1}*10,
        y error expr=\thisrowno{2}*10,
        col sep=comma
      ] {plots/uncertainty_without_qnorm/test_noise/raw_data/slab_zivid_pcd.csv};

      \addplot[
        color=Act, 
        mark=square, 
        thick
      ] table[
        x index=0,
        y expr=\thisrowno{1}*10,
        y error expr=\thisrowno{2}*10,
        col sep=comma
      ] {plots/uncertainty_without_qnorm/test_noise/raw_data/slab_zivid_pcd_uncertainty_activation.csv};

      \addplot[
        color=MC, 
        mark=diamond, 
        thick
      ] table[
        x index=0,
        y expr=\thisrowno{1}*10,
        y error expr=\thisrowno{2}*10,
        col sep=comma
      ] {plots/uncertainty_without_qnorm/test_noise/raw_data/slab_zivid_pcd_uncertainty_mcdropout.csv};

      \addplot[
        color=Grad, 
        mark=triangle, 
        thick
      ] table[
        x index=0,
        y expr=\thisrowno{1}*10,
        y error expr=\thisrowno{2}*10,
        col sep=comma
      ] {plots/uncertainty_without_qnorm/test_noise/raw_data/slab_zivid_pcd_activation_probability.csv};

    \end{axis}

    \node[anchor=north, outer sep=0, inner sep=0] at ($(Organ.north) + (0, 0.5cm)$) {\large Organ};
    \node[anchor=north, outer sep=0, inner sep=0] at ($(Cylinder.north) + (0, 0.5cm)$) {\large Cylinder};
    \node[anchor=north, outer sep=0, inner sep=0] at ($(Slab.north) + (0, 0.5cm)$) {\large Slab};

  \end{tikzpicture}

  \begin{tikzpicture}
    \begin{axis}[
      name=Organ,
      width=\myplotwidth,
      height=\myplotheight,
      ymin=-1,         %
      ymax=31,         %
      ylabel={ROI Center Error [mm]},
      ylabel style={
        at={(-0.18, 0.5)},
        font=\small,
      },
      xlabel={Dropout},
      xlabel style={
        at={(1.60, -0.15)},
        font=\small,
        anchor=north,
      },
      minor ytick={5,15,25},   %
      ymajorgrids=true,
      yminorgrids=true,
      major grid style={semithick, black!30!white, dashed},
      minor grid style={ultra thin, black!20!white, dashed},
      minor y tick style = transparent,
      x tick label style={
        font=\footnotesize,
        /pgf/number format/fixed,
        /pgf/number format/precision=5
      },
      y tick label style={
        font=\footnotesize,
      },
      scaled x ticks=false,
      axis background/.style={fill=organred},
      legend style={
        at={(1.5,1.03)},
        font=\small,
        anchor=south,
        legend columns=5
      },
    ]

      \addplot[
        color=Cent, 
        mark=star, 
        thick
      ] table[
        x index=0,
        y expr=\thisrowno{1}*10,
        y error expr=\thisrowno{2}*10,
        col sep=comma
      ] {plots/uncertainty_without_qnorm/test_drop/raw_data/organ_biopsy_zivid_pcd.csv};

      \addplot[
        color=Act, 
        mark=square, 
        thick
      ] table[
        x index=0,
        y expr=\thisrowno{1}*10,
        y error expr=\thisrowno{2}*10,
        col sep=comma
      ] {plots/uncertainty_without_qnorm/test_drop/raw_data/organ_biopsy_zivid_pcd_uncertainty_activation.csv};

      \addplot[
        color=MC, 
        mark=diamond, 
        thick
      ] table[
        x index=0,
        y expr=\thisrowno{1}*10,
        y error expr=\thisrowno{2}*10,
        col sep=comma
      ] {plots/uncertainty_without_qnorm/test_drop/raw_data/organ_biopsy_zivid_pcd_uncertainty_mcdropout.csv};

      \addplot[
        color=Grad, 
        mark=triangle, 
        thick
      ] table[
        x index=0,
        y expr=\thisrowno{1}*10,
        y error expr=\thisrowno{2}*10,
        col sep=comma
      ] {plots/uncertainty_without_qnorm/test_drop/raw_data/organ_biopsy_zivid_pcd_activation_probability.csv};

      \addplot[thick, color=OIred] coordinates {(0.5,-1) (0.5,31)};

    \end{axis}

    \begin{axis}[
      name=Cylinder,
      at=(Organ.south east),
      anchor=left of south west,
      width=\myplotwidth,
      height=\myplotheight,
      ymin=-1,         %
      ymax=31,         %
      ylabel style={
        at={(-0.08, 0.5)},
      },
      yticklabels={,,},
      minor ytick={5,15,25},
      ymajorgrids=true,
      yminorgrids=true,
      major grid style={semithick, black!30!white, dashed},
      minor grid style={ultra thin, black!20!white, dashed},
      minor y tick style = transparent,
      x tick label style={
        font=\footnotesize,
        /pgf/number format/fixed,
        /pgf/number format/precision=5
      },
      scaled x ticks=false,
      axis background/.style={fill=cylindergreen},
    ]

      \addplot[
        color=Cent, 
        mark=star, 
        thick
      ] table[
        x index=0,
        y expr=\thisrowno{1}*10,
        y error expr=\thisrowno{2}*10,
        col sep=comma
      ] {plots/uncertainty_without_qnorm/test_drop/raw_data/cylinder_v2_zivid_pcd.csv};

      \addplot[
        color=Act, 
        mark=square, 
        thick
      ] table[
        x index=0,
        y expr=\thisrowno{1}*10,
        y error expr=\thisrowno{2}*10,
        col sep=comma
      ] {plots/uncertainty_without_qnorm/test_drop/raw_data/cylinder_v2_zivid_pcd_uncertainty_activation.csv};

      \addplot[
        color=MC, 
        mark=diamond, 
        thick
      ] table[
        x index=0,
        y expr=\thisrowno{1}*10,
        y error expr=\thisrowno{2}*10,
        col sep=comma
      ] {plots/uncertainty_without_qnorm/test_drop/raw_data/cylinder_v2_zivid_pcd_uncertainty_mcdropout.csv};

      \addplot[
        color=Grad, 
        mark=triangle, 
        thick
      ] table[
        x index=0,
        y expr=\thisrowno{1}*10,
        y error expr=\thisrowno{2}*10,
        col sep=comma
      ] {plots/uncertainty_without_qnorm/test_drop/raw_data/cylinder_v2_zivid_pcd_activation_probability.csv};

      \addplot[thick, color=OIred] coordinates {(0.5,-1) (0.5,31)};

    \end{axis}

    \begin{axis}[
      name=Slab,
      at=(Cylinder.south east),
      anchor=left of south west,
      width=\myplotwidth,
      height=\myplotheight,
      ymin=-1,         %
      ymax=31,         %
      xmax=1.0,
      yticklabels={,,},
      minor ytick={5,15,25},
      ymajorgrids=true,
      yminorgrids=true,
      major grid style={semithick, black!30!white, dashed},
      minor grid style={ultra thin, black!20!white, dashed},
      minor y tick style = transparent,
      x tick label style={
        font=\footnotesize,
        /pgf/number format/fixed,
        /pgf/number format/precision=5
      },
      scaled x ticks=false,
      axis background/.style={fill=slabyellow},
    ]

      \addplot[
        color=Cent, 
        mark=star, 
        thick
      ] table[
        x index=0,
        y expr=\thisrowno{1}*10,
        y error expr=\thisrowno{2}*10,
        col sep=comma
      ] {plots/uncertainty_without_qnorm/test_drop/raw_data/slab_zivid_pcd.csv};

      \addplot[
        color=Act, 
        mark=square, 
        thick
      ] table[
        x index=0,
        y expr=\thisrowno{1}*10,
        y error expr=\thisrowno{2}*10,
        col sep=comma
      ] {plots/uncertainty_without_qnorm/test_drop/raw_data/slab_zivid_pcd_uncertainty_activation.csv};

      \addplot[
        color=MC, 
        mark=diamond, 
        thick
      ] table[
        x index=0,
        y expr=\thisrowno{1}*10,
        y error expr=\thisrowno{2}*10,
        col sep=comma
      ] {plots/uncertainty_without_qnorm/test_drop/raw_data/slab_zivid_pcd_uncertainty_mcdropout.csv};

      \addplot[
        color=Grad, 
        mark=triangle, 
        thick
      ] table[
        x index=0,
        y expr=\thisrowno{1}*10,
        y error expr=\thisrowno{2}*10,
        col sep=comma
      ] {plots/uncertainty_without_qnorm/test_drop/raw_data/slab_zivid_pcd_activation_probability.csv};

      \addplot[thick, color=OIred] coordinates {(0.5,-1) (0.5,31)};

    \end{axis}

  \end{tikzpicture}

  \caption{The effect of adding noise (as a fraction of the scene size) and dropping a fraction of points from the input point cloud on the accuracy of estimating the center of the Regions of Interest (ROIs) is evaluated. Different centroid estimation methods are considered: geometric centroid (GC), activation-based (Act.-UWC), and Monte-Carlo Dropout (MCD-UWC) based uncertainty-weighted centroid estimation, and finally softmax probability-weighted (SPWC) centroid estimation. The vertical line at $\text{Dropout} = 0.5$ indicates the maximum proportion of points randomly dropped during training.}
  \label{plot:local_uncertainty}
\end{figure}

    \fi

        \paragraph{Local Per-Point Uncertainty}
        Using the local per-point uncertainty and probability values for weighting the centroid estimation results in the same or decreased accuracy.
        \Cref{plot:local_uncertainty} shows that \gls{uwc} with activation entropy and \gls{spwc} perform almost identically to the geometric centroid of an \gls{roi}.
        Using \gls{mcd} uncertainty values results in a strong error increase as noise levels increase or more input point are dropped.
        The effect of dropping up to $60\%$ of input points has no effect on the centroid error.
        This roughly coincides with the number of points dropped during training ($50\%$).
        We did not perform experiments for \textit{Slab} with noise levels above $0.01$ as the \gls{on} fails to consistently reconstruct all \glspl{roi}.
    \ifrenderfigures
        \input{figures/uncertainty_visual_examples}
    \fi

        The visualization in \Cref{fig:uncertainty_visual_examples} shows that uncertainty is high close to the decision boundaries between segments of the object or between the object and the outside.
        When adding noise (noise level $0.03$) the decision boundaries become more spread out, indicating \gls{methodname} is less confident about the true location of the decision boundaries.
        To quantify this spreading out, we can consider the aggregated uncertainty of the reconstruction.
        
        \paragraph{Aggregated Uncertainty}
    \ifrenderfigures

\definecolor{colorcylinder}{rgb}{0.1,0.5,0.1}
\definecolor{colororgan}{rgb}{0.5,0.1,0.1}
\definecolor{colorslab}{rgb}{0.5,0.5,0.1}

\definecolor{OIblack}{RGB}{0, 0, 0}
\definecolor{OIgreen}{RGB}{0, 158, 115}
\definecolor{OIblue}{RGB}{0, 114, 178}
\definecolor{OIlightblue}{RGB}{86, 180, 233}
\definecolor{OIyellow}{RGB}{240, 228, 66}
\definecolor{OIorange}{RGB}{230, 159, 0}
\definecolor{OIred}{RGB}{213, 94, 0}
\definecolor{OIpink}{RGB}{204, 121, 167}

\colorlet{Cent}{OIblue}
\colorlet{Act}{OIgreen}
\colorlet{MC}{OIorange}
\colorlet{Grad}{OIpink}

\definecolor{organred}{HTML}{FFF0F0}
\definecolor{cylindergreen}{HTML}{F0FFF0}
\definecolor{slabyellow}{HTML}{FFFFF0}

\def\myplotwidth{0.45\columnwidth}
\def\myplotheight{0.45\columnwidth}

\begin{figure}
\centering
\vspace{1.5mm}
\begin{tikzpicture}
\begin{axis}[
    name=Organ,
    width=\myplotwidth,
    height=\myplotheight,
    ymin=0.0,
    ymax=0.3,
    legend style={
        at={(0.0,1.25),
        font=\small,
        },
    anchor=south west,
    legend columns=4},
    xlabel={Noise},
    xlabel style={
        at={(1.60, -0.15)},
        font=\small,
        anchor=north,
    },
    ylabel={Aggr. Uncertainty},
    ylabel style={
        at={(-0.18, 0.5)},
        font=\small,
    },
    minor ytick={0,0.05,0.15,0.25},
    ymajorgrids=true,
    yminorgrids=true,
    major grid style={semithick, black!30!white, dashed},
    minor grid style={ultra thin, black!20!white, dashed},
    minor y tick style = transparent,
    x tick label style={
        font=\footnotesize,
        /pgf/number format/fixed,
        /pgf/number format/precision=5
    },
    scaled x ticks=false,
    y tick label style={
        font=\footnotesize,
    },
    axis background/.style={fill=organred},
]

\addplot[color=Act, mark=square, thick] table[x index=0, y index=1, y error index=2, col sep=comma] {plots/uncertainty_without_qnorm/global_uncertainty_noise/data_Organ_Biopsy.csv};
\addlegendentry{Act.}

\addplot[color=MC, mark=diamond, thick] table[x index=0, y index=2, y error index=2, col sep=comma] {plots/uncertainty_without_qnorm/global_uncertainty_noise/data_Organ_Biopsy.csv};
\addlegendentry{MCD}

\end{axis}

\begin{axis}[
    name=Cylinder,
    at=(Organ.south east),
    anchor=left of south west,
    width=\myplotwidth,
    height=\myplotheight,
    ymin=0.0,
    ymax=0.3,
    ylabel style={
        at={(-0.08, 0.5)},
    },
    yticklabels={,,},
    minor ytick={0,0.05,0.15,0.25},
    ymajorgrids=true,
    yminorgrids=true,
    major grid style={semithick, black!30!white, dashed},
    minor grid style={ultra thin, black!20!white, dashed},
    minor y tick style = transparent,
    x tick label style={
        font=\footnotesize,
        /pgf/number format/fixed,
        /pgf/number format/precision=5
    },
    scaled x ticks=false,
    axis background/.style={fill=cylindergreen},
]

\addplot[color=Act, mark=square, thick] table[x index=0, y index=1, y error index=2, col sep=comma] {plots/uncertainty_without_qnorm/global_uncertainty_noise/data_Cylinder.csv};

\addplot[color=MC, mark=diamond, thick] table[x index=0, y index=2, y error index=2, col sep=comma] {plots/uncertainty_without_qnorm/global_uncertainty_noise/data_Cylinder.csv};

\end{axis}

\begin{axis}[
    name=Slab,
    at=(Cylinder.south east),
    anchor=left of south west,
    width=\myplotwidth,
    height=\myplotheight,
    ymin=0.0,
    ymax=0.3,
    xtick distance=0.02,
    ylabel style={
        at={(-0.08, 0.5)},
    },
    yticklabels={,,},
    minor ytick={0,0.05,0.15,0.25},
    ymajorgrids=true,
    yminorgrids=true,
    major grid style={semithick, black!30!white, dashed},
    minor grid style={ultra thin, black!20!white, dashed},
    minor y tick style = transparent,
    x tick label style={
        font=\footnotesize,
        /pgf/number format/fixed,
        /pgf/number format/precision=5
    },
    scaled x ticks=false,
    axis background/.style={fill=slabyellow},
]

\addplot[color=Act, mark=square, thick] table[x index=0, y index=1, y error index=2, col sep=comma] {plots/uncertainty_without_qnorm/global_uncertainty_noise/data_Slab.csv};

\addplot[color=MC, mark=diamond, thick] table[x index=0, y index=2, y error index=2, col sep=comma] {plots/uncertainty_without_qnorm/global_uncertainty_noise/data_Slab.csv};
\end{axis}

\node[anchor=north, outer sep=0, inner sep=0] at ($(Organ.north) + (0, 0.5cm)$) {\large Organ};
\node[anchor=north, outer sep=0, inner sep=0] at ($(Cylinder.north) + (0, 0.5cm)$) {\large Cylinder};
\node[anchor=north, outer sep=0, inner sep=0] at ($(Slab.north) + (0, 0.5cm)$) {\large Slab};

\end{tikzpicture}

 \vspace{0.3cm}

\begin{tikzpicture}
\begin{axis}[
    name=Organ,
    width=\myplotwidth,
    height=\myplotheight,
    ymin=0.0,
    ymax=0.5,
    legend style={
        at={(1.5,0.9),
        font=\small,
    },
    anchor=south,
    legend columns=4},
    xlabel={Dropout},
    xlabel style={
        at={(1.60, -0.15)},
        font=\small,
        anchor=north,
    },
    ylabel={Aggr. Uncertainty},
    ylabel style={
        at={(-0.18, 0.5)},
        font=\small,
    },
    minor ytick={0,0.1,0.3},
    ymajorgrids=true,
    yminorgrids=true,
    major grid style={semithick, black!30!white, dashed},
    minor grid style={ultra thin, black!20!white, dashed},
    minor y tick style = transparent,
    x tick label style={
        font=\footnotesize,
        /pgf/number format/fixed,
        /pgf/number format/precision=5
    },
    scaled x ticks=false,
    y tick label style={
        font=\footnotesize,
    },
    axis background/.style={fill=organred},
]

\addplot[color=Act, mark=square, thick] table[x index=0, y index=1, y error index=2, col sep=comma] {plots/uncertainty_without_qnorm/global_uncertainty_point_drop/data_organ_biopsy.csv};

\addplot[color=MC, mark=diamond, thick] table[x index=0, y index=2, y error index=2, col sep=comma] {plots/uncertainty_without_qnorm/global_uncertainty_point_drop/data_organ_biopsy.csv};
\addplot[thick, color=OIred] coordinates {(0.5,-0.1) (0.5,0.75)};

\end{axis}

\begin{axis}[
    name=Cylinder,
    at=(Organ.south east),
    anchor=left of south west,
    width=\myplotwidth,
    height=\myplotheight,
    ymin=0.0,
    ymax=0.5,
    ylabel style={
        at={(-0.08, 0.5)},
    },
    yticklabels={,,},
    minor ytick={0,0.1,0.3},
    ymajorgrids=true,
    yminorgrids=true,
    major grid style={semithick, black!30!white, dashed},
    minor grid style={ultra thin, black!20!white, dashed},
    minor y tick style = transparent,
    x tick label style={
        font=\footnotesize,
        /pgf/number format/fixed,
        /pgf/number format/precision=5
    },
    scaled x ticks=false,    
    axis background/.style={fill=cylindergreen},
]

\addplot[color=Act, mark=square, thick] table[x index=0, y index=1, y error index=2, col sep=comma] {plots/uncertainty_without_qnorm/global_uncertainty_point_drop/data_cylinder.csv};

\addplot[color=MC, mark=diamond, thick] table[x index=0, y index=2, y error index=2, col sep=comma] {plots/uncertainty_without_qnorm/global_uncertainty_point_drop/data_cylinder.csv};
\addplot[thick, color=OIred] coordinates {(0.5,-0.1) (0.5,0.75)};

\end{axis}

\begin{axis}[
    name=Slab,
    at=(Cylinder.south east),
    anchor=left of south west,
    width=\myplotwidth,
    height=\myplotheight,
    ymin=0.0,
    ymax=0.5,
    xtick distance=0.5,
    ylabel style={
        at={(-0.08, 0.5)},
    },
    yticklabels={,,},
    minor ytick={0,0.1,0.3},
    ymajorgrids=true,
    yminorgrids=true,
    major grid style={semithick, black!30!white, dashed},
    minor grid style={ultra thin, black!20!white, dashed},
    minor y tick style = transparent,
    x tick label style={
        font=\footnotesize,
        /pgf/number format/fixed,
        /pgf/number format/precision=5
    },
    scaled x ticks=false,
    axis background/.style={fill=slabyellow},
]

\addplot[color=Act, mark=square, thick] table[x index=0, y index=1, y error index=2, col sep=comma] {plots/uncertainty_without_qnorm/global_uncertainty_point_drop/data_slab.csv};

\addplot[color=MC, mark=diamond, thick] table[x index=0, y index=2, y error index=2, col sep=comma] {plots/uncertainty_without_qnorm/global_uncertainty_point_drop/data_slab.csv};
\addplot[thick, color=OIred] coordinates {(0.5,-0.1) (0.5,0.75)};

\end{axis}

\end{tikzpicture}

\caption{The aggregated uncertainty is evaluated as noise is added to the input point cloud (as a fraction of the object size) and as points are dropped from the input point cloud. The evaluation compares different centroid estimation methods: Activation-based (Act.) and Monte-Carlo Dropout (MCD) based uncertainty. The vertical line at $\text{Dropout} = 0.5$ indicates the maximum proportion of points that were randomly dropped during training.}
\label{plot:global_uncertainty}
\end{figure}
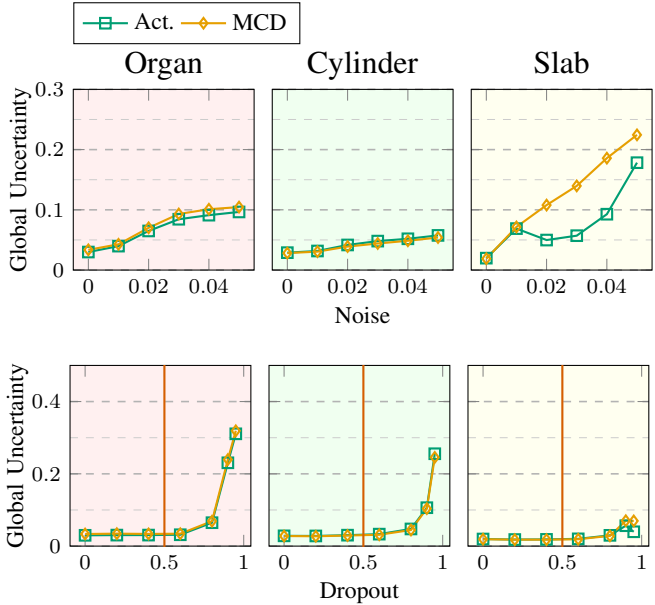

    \fi

        Both activation entropy and \gls{mcd}-based uncertainty values, when aggregated, can be used to obtain a single uncertainty value.
        \Cref{plot:global_uncertainty} shows that, with an exception for \textit{Slab}, both approaches provide strictly monotonically increasing aggregated uncertainty values as noise or input point drop level increase.
        The noise level of $0.01$, at which the activation entropy-based aggregated uncertainty for \textit{Slab} suddenly drops, coincides with the point where the \gls{on} fails to correctly reconstruct all \glspl{roi}.

\begin{figure}[t]
    \centering
    \begin{tikzpicture}
        \node[anchor=south west] (a) at (0,0) {\includegraphics[width=0.3\columnwidth]
        {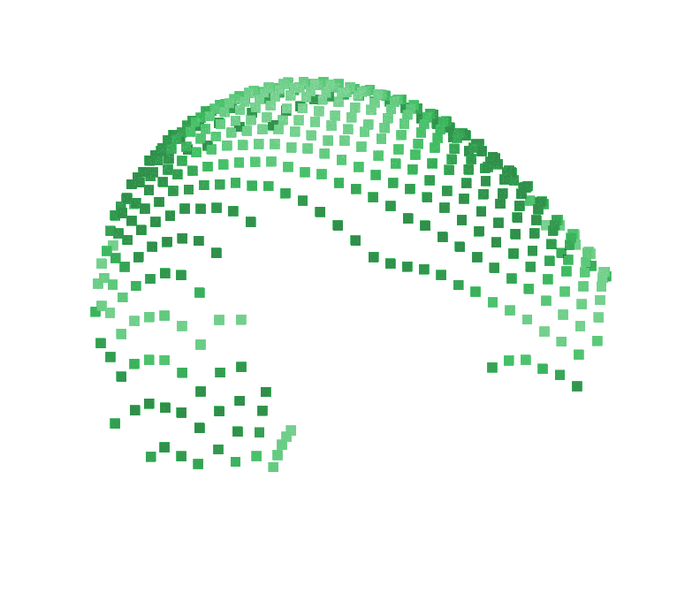}};
        \node[below=0.3cm of a.south, anchor=center] {$\text{AU} = 0.052$};
        \node[above left=-0.3cm and -0.1cm of a.north west, anchor=north west] {(a)};
        
        \node[anchor=south west] (b) at (3,0) {\includegraphics[width=0.3\columnwidth]{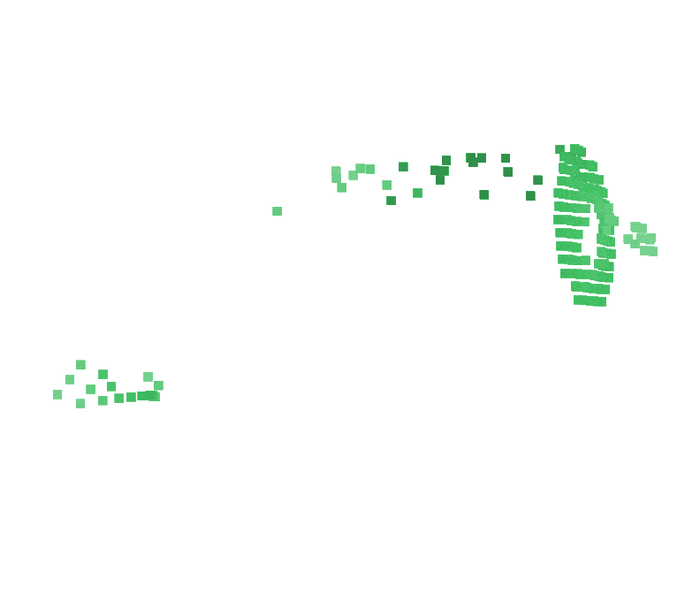}};
        \node[below=0.3cm of b.south, anchor=center] {$\text{AU} = 0.067$};
        \node[above left=-0.3cm and -0.3cm of b.north west, anchor=north west] {(b)};
        
        \node[anchor=south west] (c) at (6,0) {\includegraphics[width=0.3\columnwidth]{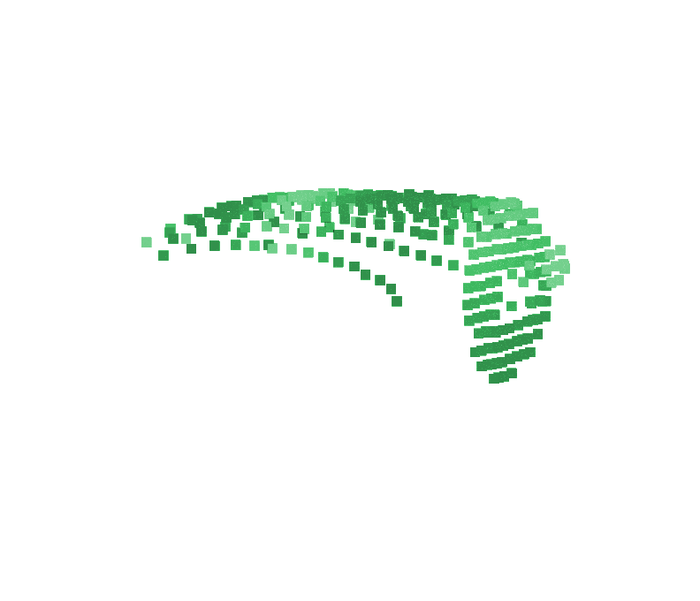}};
        \node[below=0.3cm of c.south, anchor=center] {$\text{AU} = 0.100$};
        \node[above left=-0.3cm and -0.3cm of c.north west, anchor=north west] {(c)};
    \end{tikzpicture}
    \caption{
        Examples of input point clouds $P$ of the \textit{Cylinder} with activation-based Aggregated Uncertainty (AU) outside the range $0.029 \pm 0.014$ (mean $\pm$ standard deviation).  
        (a) The \textit{Cylinder} is bent into an inverted U-shape, leading to ambiguity in labeling the outer Regions of Interest (ROIs).
        (b) A point cloud with large missing regions, caused by self-occlusion, but still capturing both ends.  
        (c) A point cloud containing only the top part of the \textit{Cylinder}, with the bottom half not visible at all.
    }
    \label{fig:inputs_with_high_uncertainty}
\end{figure}

        Additionally, we provide examples of inputs that result in high aggregated uncertainty values for the \textit{Cylinder} without added noise and without input point drop, see \Cref{fig:inputs_with_high_uncertainty}.

    \subsection{Explainability}
    \ifrenderfigures
        \input{figures/explainability_examples}
    \fi

        We qualitatively present explainability point clouds obtained from each object in \Cref{fig:explainability_examples}.
        \gls{methodname} captures features that are important to understanding the deformations of each deformable object.
        For the \textit{Cylinder}, the attachment point at the base and the regions of interaction are clearly highlighted.
        The two interaction regions at the tips of the \textit{Slab} are also clearly highlighted.
        For the \textit{Organ}, the bump at the top is highlighted, identifying it as a strong feature which also indicates the position of the embedded middle \gls{roi}.

    \subsection{Robot Calibration}
        Our robot calibration procedure described in \Cref{sec:robotic_calibration} decreases the mean positioning error from \SI{5.466}{\mm} to \SI{0.956}{\mm}.
        The maximum error decreases from \SI{8.752}{\mm} to \SI{2.699}{\mm}.
        The ablative experiments described in \Cref{sec:experiments_robot_calibration} affect the mean positioning error as follows:

        \begin{tabular}{l l}
            & \\[-1.5ex]
            \textbf{$\bullet$ ($\Delta\theta$-Only Calibration)}: & +\SI{1.677}{\mm}, \\[0.5ex]
            \textbf{$\bullet$ (Fixed $T_{\text{base}}$)}: & +\SI{0.035}{\mm}, \\[0.5ex]
            \textbf{$\bullet$ (Dynamic Data)}: & +\SI{+0.161}{\mm}, and  \\[0.5ex]
            \textbf{$\bullet$ (Single Joint Movement)}: & +\SI{+0.129}{\mm}.\\[1ex]
        \end{tabular}

        The largest increase in positioning error is a result of limiting the optimization to only $\Delta\theta$.
        This shows the importance of calibrating the complete set of \gls{dh} parameters in contrast to calibrating only the joint zero positions $\theta$.
        The low impact of calibrating $T_\text{base}$ indicates a good initial estimate for the global position of the robot base in relation to the world reference frame.
        The use of dynamically collected data (\ie using samples $(p_{\text{tracked}}, q_{\text{tracked}}, \xi)$ collected during the movement between end-points) results in a minor decrease in accuracy.
        The specific choice of motion to generate the training data has only minor impact on the total accuracy.

\begin{table}[t]
    \centering
    \renewcommand{\arraystretch}{1.2}
    \setlength{\tabcolsep}{6pt}
    \caption{Mean centroid error (mm) for V2S and LUDO after $100$ epochs of training across the three deformable objects. The \textcolor{OIred}{\textbf{FO}} and \textcolor{OIgreen}{\textbf{FT}} stand for \textcolor{OIred}{Full Observation} and \textcolor{OIgreen}{Fine-Tuned}. $\text{LUDO}^*$ was trained for 600 epochs. Rigid Alignment represents the centroid error of the rigidly aligned data given to V2S as input. The mean $\pm$ standard deviation values are computed over $64$ different deformations for each object.}
    \begin{tabular}{lccc}
        \toprule
        Method & \textbf{Organ} & \textbf{Cylinder} & \textbf{Slab} \\
        \midrule
        Rigid Alignment     & $12.5 \pm 8.0$ & $44.7 \pm 14.8$ & $24.3 \pm 8.2$ \\
        V2S                 & $2.7 \pm 1.7$  & $18.8 \pm 7.7$  & $13.9 \pm 6.9$ \\
        V2S+\textcolor{OIred}{\textbf{FO}}   & $2.4 \pm 1.6$  & $18.4 \pm 6.9$  & $9.9 \pm 3.7$  \\
        V2S+\textcolor{OIgreen}{\textbf{FT}} & $1.9 \pm 1.3$  & $3.8 \pm 1.8$   & $5.1 \pm 2.6$  \\
        V2S+\textcolor{OIgreen}{\textbf{FT}}+\textcolor{OIred}{\textbf{FO}} & $1.4 \pm 0.8$  & $2.5 \pm 1.3$   & $3.7 \pm 2.0$  \\
        LUDO (Ours)        & $\mathbf{1.4 \pm 0.5}$  & $\mathbf{1.9 \pm 0.7}$  & $\mathbf{1.9 \pm 1.0}$  \\
        \midrule
        $\text{LUDO}^*$ (Ours) & $\mathbf{0.7 \pm 0.3}$  & $\mathbf{0.7 \pm 0.4}$  & $\mathbf{1.2 \pm 0.6}$  \\
        \bottomrule
    \end{tabular}
    \label{tab:v2s_vs_ludo}
\end{table}

    \subsection{Comparison to Baseline}
        The mean centroid errors of \gls{methodname} and \gls{v2s} are presented in \Cref{tab:v2s_vs_ludo}.
        After training for 100 epochs, \gls{methodname} outperforms \gls{v2s} with fine-tuning (FT), full observations (FO) and an optimal alignment (V2S+FT+FO) across all three evaluated objects.
        We found that the registration quality of \gls{v2s} plateaus after approximately $70$ to $90$ epochs of fine-tuning, which aligns with the number of epochs used in the original work.
        \gls{methodname}'s accuracy does not plateau after $100$ epochs, the centroid error is reduced further through additional epochs.
        On the smaller training dataset used for the baseline comparison, \gls{methodname} requires between $20$ and $40$ seconds per training epoch depending on the scene.
        \gls{v2s} requires an average of $14$ minutes per epoch.
        Therefore, \gls{methodname} can be trained between $21$ and $42$ times faster on the same hardware.
        This large difference in training time also correlates with the raw file size of the training data.
        An average \gls{methodname} training dataset is approximately $10$~gigabytes in size, whereas the volumetric representation used by \gls{v2s} results in training datasets of approximately $160$~gigabytes for each object.
        For \gls{v2s}, the training time is in strong contrast to the inference time.
        \gls{v2s} requires \SI{5}{ms} whereas \gls{methodname} requires approximately \SI{20}{ms}.

    \subsection{Real-World Robotic Puncturing}
        For the real-world robotic puncturing we use \gls{methodname} with geometric centroids and $40\,000$ query points.
        The success rate (\ie, hitting the \glspl{roi} with the needle) for the \textit{Organ} and \textit{Slab} was $100\%$ for all \glspl{roi}.
        For \textit{Cylinder} the success rate was $96.67\%$, with a single failure for the centered \gls{roi}.
        Overall, we attempted $90$ \gls{roi} punctures and achieved a success rate of $98.9\%$.
        The single failure case is discussed in \Cref{sec:puncture_performance}.

    \section{DISCUSSION}
    \label{sec:discussion}
        \subsection{Inference Time}
            The inference time of \gls{methodname} is an important factor, in particular for real-time applications like robotic surgeries.
            In our experiments, we find that using $40\,000$ query points achieves a good balance between speed and accuracy, with inference times consistently below \SI{30}{\ms}.
            This performance ensures that the system can provide low-latency feedback necessary for dynamic environments where deformations can occur rapidly, such as surgical procedures.
            \gls{methodname} has a constant inference time, regardless of the complexity of the deformation. 
            The only variables that affects inference time are the number of points in the observation, see difference between Slab and Cylinder, as well as the number of points used to query the occupancy network.
            Both of these variables can be adjusted online to ensure a constant inference time.
            For example by randomly sub-sampling the point cloud observation, which \gls{methodname} is robust to as shown in \Cref{plot:global_uncertainty}.

        \subsection{Uncertainty Estimation}
            Both \gls{mcd} and activation based uncertainty estimation provide very similar outputs.
            Contrary to our initial expectations, we found that using the uncertainty values provided by both methods for a weighted \gls{roi} centroid estimation decreased accuracy.
            Especially the use of \gls{mcd} weighting resulted in a strong degradation of accuracy.
            The use of \gls{spwc} seems to also not provide a benefit over using the geometric centroid.
            Nevertheless, both uncertainty approaches can be used to estimate a aggregated uncertainty.
            Intuitively, as the decision boundaries spread out with increased noise, see \Cref{fig:uncertainty_visual_examples}, the aggregated uncertainty values increase, see \Cref{plot:global_uncertainty}.
            We found that when \gls{methodname} becomes highly uncertain, reaching a point where it fails to reconstruct \glspl{roi}, the activation entropy-based aggregated uncertainty estimation unexpectedly exhibits a sudden drop, contrary to the anticipated continuous increase.
            Activation entropy-based uncertainty estimation uses a single forward pass of the model, and cannot distinguish between confident correct and confident incorrect predictions.
            In contrast, \gls{mcd}-based uncertainty estimation aggregates the predictions from multiple forward passes.
            The disagreements between confident incorrect predictions increase the entropy of the aggregated prediction.
            Increased entropy reveals uncertainty even when individual models are confident in their incorrect predictions.
            Thus, \gls{mcd}-based aggregated uncertainty is better suited for high-risk applications.
            Nevertheless, the additional computation cost of \gls{mcd}-based uncertainty needs to be taken into account.
            The inference time increases linearly with the number of inferences performed with different dropout seeds.
            Finally, the amount of increase in aggregated uncertainty is scene dependent.
            Aggregated uncertainty normalization that is object-independent remains an open problem for future work.
            We currently use a uniform sampling strategy to query the neural occupancy function within a bounding box, which is first fitted to the normalized observation and then refined to focus on regions likely containing an object.
            Consequently, the sampling distribution varies and cannot be considered independent and identically distributed.
            For example, a tighter fitting bounding box, as opposed to a looser one, results in fewer sampled points outside the object, with those points also being closer to its surface.
            Since points far from the decision boundary typically exhibit low uncertainty, a tighter bounding box tends to result in higher aggregated uncertainty.
            Nonetheless, our aggregated uncertainty provides a practical indicator for model confidence.
            If the average entropy increases (\ie, boundaries become more diffuse), it signals lower overall confidence.

            \noindent\textbf{Detecting Ambiguities:} For strongly deforming objects or objects with symmetries, estimating structures using single-view perspectives can become ambiguous.
            During training, we found that in instances where the \textit{Cylinder} forms an upside-down U (and where the small extrusion at the top is not visible), \gls{methodname} will mix the labels for the outer \glspl{roi}.
            As such ambiguities can not be resolved without additional knowledge, it is important to detect ambiguous observations.
            \gls{methodname} can detect ambiguities through the aggregated uncertainty, which intuitively increases with increasing ambiguity, see \Cref{fig:inputs_with_high_uncertainty}.

        \subsection{Explainability}
            We found that our masking-based approach results in interpretable results for deformable object understanding.
            Areas that provide important information about the object position or deformations are clearly highlighted.
            Providing information on important features can aid in human-robot collaboration tasks, by providing a human collaborator with reasoning for autonomous decisions.
            Potential benefits or risks of such explainability visualizations should be investigated in the future.

        \subsection{Robot Calibration}
            Our calibration routine improved the manufacturer provided calibration from a mean positioning error of \SI{5.466}{\mm} to \SI{0.956}{\mm}, decreasing the maximum positioning error from \SI{8.752}{\mm} to \SI{2.699}{\mm}.
            The optimized calibration had a direct impact on the puncturing task, ensuring that the robot could hit the target \gls{roi}'s center provided by the \gls{on}.
            We found that without the calibration, the task of accurately puncturing the \glspl{roi} (diameter of \SI{17}{\mm}) was not feasible.
            Even with optimal predictions from \gls{methodname}, an inaccurate robotic system could not perform the puncturing task, as it requires precise execution of the planned trajectory.
            We performed several ablations to the calibration routine and found that limiting the calibration to only $\Delta\theta$ results in the largest decrease in accuracy.
            Still, the results indicate that the error is only partially caused by incorrectly calibrated zero positions of the rotational joints in the kinematic chain.

        \subsection{Comparison to Baseline}
        In our experiments, \gls{methodname} consistently outperforms all configurations of \gls{v2s} in accurately localizing internal structures of deformable objects.
        This is despite \gls{v2s} being provided with an optimal rigid alignment, fine-tuning, and full surface observations (V2S+FT+FO).
        Not only does \gls{methodname} achieve lower centroid errors, but it also trains orders of magnitude faster on the same hardware.
        This is an important property when considering patient-specific scenarios.
        Similarly, the datasets used for \gls{methodname} are also orders of magnitude smaller.
        This is due to the memory-inefficient 3D volumetric representations used by \gls{v2s}.

        Despite the benefits of \gls{methodname}, the baseline method \gls{v2s} has advantages.
        Its conditioning approach, which also uses a volumetric representation of the prior model as input, allows \gls{v2s} to register new deformable objects that are similarly shaped to those seen during training.
        When tested on the \textit{Organ} object, \gls{v2s} achieves a centroid localization error of $2.7\pm1.7$~mm (mean $\pm$ standard deviation) without additional fine-tuning, likely because the \textit{Organ} has a shape similar to objects in \gls{v2s}'s original training data.
        By contrast, \gls{methodname} is specialized to the single prior model used during training.
        Moreover, \gls{v2s} offers a low inference time of around \SI{5}{\ms}, compared to about \SI{20}{\ms} with \gls{methodname}.
        We want to highlight that this time measurement assumes an optimal initial alignment has been found for \gls{v2s} and all observations are already converted to a volumetric representation.

        \subsection{Real-World Robotic Puncturing}
            \label{sec:puncture_performance}
    \ifrenderfigures

\newdimen\OccupancyNetworkX
\newdimen\OccupancyPointCloudY

\definecolor{OIblack}{RGB}{0, 0, 0}
\definecolor{OIgreen}{RGB}{0, 158, 115}
\definecolor{OIblue}{RGB}{0, 114, 178}
\definecolor{OIlightblue}{RGB}{86, 180, 233}
\definecolor{OIyellow}{RGB}{240, 228, 66}
\definecolor{OIorange}{RGB}{230, 159, 0}
\definecolor{OIred}{RGB}{213, 94, 0}
\definecolor{OIpink}{RGB}{204, 121, 167}

\begin{figure}
    \vspace{0.15cm} %
    \centering
    \begin{tikzpicture}[node distance=0.2cm, auto]
        \node[anchor=south west,inner sep=0] (image) at (0,0) {\includegraphics[width=\columnwidth]{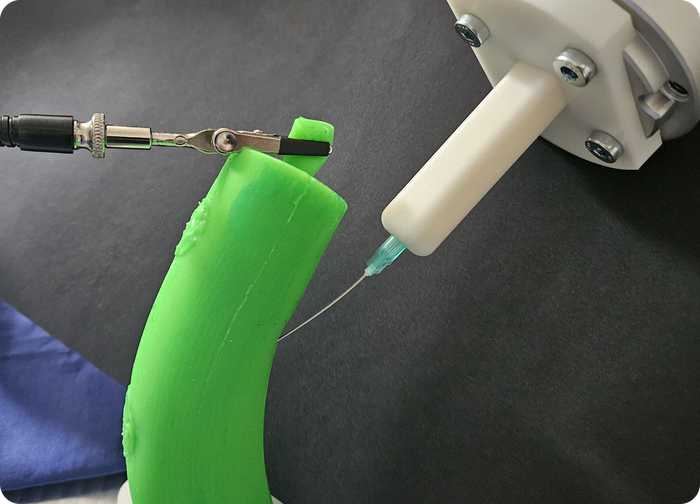}};

        \begin{scope}[x={(image.south east)}, y={(image.north west)}, yscale=1.3887] %
            \draw[OIblack, dash pattern={on 9pt off 4pt}, line width=1mm] (0.28,0.1) circle[radius=0.05]; %

            \draw[white, <-] (0.48,0.25) -- (0.78,0.25) node[pos=1, above] {Bent Needle};

            \draw[white, <-] (0.43,0.175) -- (0.73,0.175) node[pos=1, above] {Intended Trajectory};
            
            \draw[white, <-] (0.38,0.1) -- (0.68,0.1) node[pos=1, above] {Target ROI};

            \draw[OIred, opacity=0.7, dashed, ultra thick] (0.31,0.09) -- (0.78,0.6);

        \end{scope}
    \end{tikzpicture}
    \caption{Failure case during robotic execution in the Cylinder scene. The Region of Interest (ROI) was missed by the robotically controlled needle. The high friction inside of the soft silicone phantom caused the needle to bend and miss the ROI. The \textit{Intended Trajectory} is shown as a dashed line.}
    \label{fig:failure_case}
\end{figure}

    \fi

            Our approach achieved a near-perfect success rate, with $100\%$ accuracy on two of the phantoms and $96.67\%$ on the third.
            The system effectively handled various strong deformation states. 
            The single failure was due to needle bending, likely caused by friction between the needle and the silicone phantom material, see \Cref{fig:failure_case}.
            To address friction-based failure cases in a phantom setup, better entry angles or materials should be considered.
            Online replanning could also compensate for deformations during execution and will be explored in future work.  
            Overall, the experiments confirmed that the proposed method is effective for autonomous robotic puncturing.
        \section{Limitations}
            Our approach has multiple limitations that will be addressed in future work.
            \noindent\textbf{Trajectory Updates:} We do not use live updates of the reconstruction during the puncturing process.
            But due to the \highly deformable nature of our phantoms, there are noticeable deformations that occur during puncturing.
            For future applications, updating the trajectory to account for these deformations will be crucial to increase the accuracy and safety.
            \noindent\textbf{Interaction Points:} Our approach requires the generation of training data using a \gls{fem} simulation.
            Although the simulation is physically realistic, we believe a pipeline that does not require the definition of interaction points will be vital for future work.
            \noindent\textbf{Single-Patch Explainability:}
            Our current masking-based explainability method only addresses ambiguities coming from the removal of a single circular patch.  
            Consider a rigid cylinder of known length $L$ and fixed radius.
            If the top of the cylinder is masked in the observation, the visible bottom and side allow inference of the full 3D pose, as $L$ determines the missing top's location.  
            However, if both the top and bottom are masked, the visible side alone cannot resolve the cylinder’s position along its length.
            It could be shifted up or down while still matching the visible data.  
            Addressing such interdependencies is left for future work.
            \noindent\textbf{Per-ROI Explainability:}
            \gls{methodname} reconstructs the entire deformable object, including its internal structures, leading to entangled input contributions across segments.
            We can provide per-ROI explainability by computing the $S_i$ only for the respective class and omitting the other classes.
            Still, the explanation remains influenced by the holistic reconstruction.
            A more targeted approach, where the network is trained to reconstruct only the \gls{roi} given the observation, could improve the interpretability of per-segment input contributions.
            This approach comes with the limitation that the structural information for the encompassing object is not available for trajectory planning.
            Disentangling explanations for multi-part object reconstruction remains an open task for future work.
            \noindent\textbf{Normalization of Aggregated Uncertainty:} The aggregated uncertainty values depend on the object, making it unfeasible to establish a single overall uncertainty threshold for determining when the output is unsafe for use.
            An additional neural network could be used to learn a scene-specific uncertainty normalization.
            Additionally, strategies should be explored to ensure that no bias is introduced due to the query point sampling strategy. 
            \noindent\textbf{Safety:} We wish to emphasize that although \gls{methodname} provides explainability and uncertainty estimates, we do not evaluate the overall safety of \gls{methodname}.
            \noindent\textbf{Background Segmentation:} Our approach requires the input point cloud to contain only the deformable object of interest.
            Therefore, a segmentation of the point cloud is needed.
            As many depth sensors also provide an RGB image, machine learning-based image segmentation approaches should be investigated for background segmentation in future work.
            \noindent\textbf{Depth Camera:} The physical dimensions of our depth sensor constrain its practical use in surgical settings.
            In minimally invasive surgery, a more compact depth sensor would be necessary.
            Additionally, the Zivid One+ M achieves high-precision depth estimations through structured light projection.
            However, projecting structured patterns onto an intraoperative scene could interfere with the surgical workflow.
            Consequently, the quality of the point cloud observations in our evaluation is likely superior to what can be obtained in real surgical environments, where stereo-based microscopic or endoscopic systems are more commonly used.
            Nevertheless, we currently observe a fast development of accurate machine learning-based monocular and stereo depth estimation methods.
            Such approaches are quickly becoming competitive and should be considered in future work.
            \noindent\textbf{Small Segments:} As we use dense output point clouds to determine target positions, interacting with very small object segments can be challenging, as they are less likely to be queried using random query points.
            An approach that better queries areas where structures are likely to be, such as multi-staged hierarchical querying, could improve the inference of very small structures.
            \noindent\textbf{Controlled Environment:}
            The experimental setup involved a controlled environment, primarily due to precision requirements of the robotic system.
            The camera and robot are rigidly mounted to the table to ensure accurate positioning, and the organ phantoms are secured to prevent movement during puncturing.
            This setup does not fully capture real-world conditions with complex backgrounds and soft tissue that moves during puncturing.
            Addressing these constraints in a dynamic setting will require online re-planning and a robust background segmentation.

\section{CONCLUSION}
\label{sec:conclusion}
    This work introduces \gls{methodname}, a method for reconstructing and targeting internal structures of deformable objects for robotic applications.
    \gls{methodname} uses neural occupancy networks to infer the full state of deformable objects and their internal structures.
    The fast inference times of below \SI{30}{\ms} make this approach highly applicable in interactive real-world applications.
    In contrast to previous approaches for deformable object reconstruction, we generate our training data using \gls{sofa}, a state-of-the-art \gls{fem} simulation framework.
    We investigate methods and give experimental insights into local and aggregated uncertainty estimation for occupancy learning.
    Additionally, we introduce a masking based explainability approach for 3D reconstructions from point clouds.
    We validate our approach in robotic real-world experiments, where we puncture \glspl{roi} in three \highly deformable objects with a total success rate of $98.9\%$.
    One notable failure case, where the robotic needle missed the \gls{roi} due to bending, emphasizes the importance of low-latency trajectory updates during deformable object interactions.
    Such online trajectory updates will be the focus of future investigations.
    We demonstrated \gls{methodname}'s advantages compared to \gls{v2s}, despite providing the baseline with numerous advantages during the evaluation.
    \gls{methodname} is an important step towards practical and safe robotic interaction with \highly deformable objects, as it not only provides rapid structural information, including internal structures hidden from visual observation, but also introduces preliminary methods for quantifying its own uncertainty.
    \gls{methodname} eliminates the need for traditional registration methods and sets a new foundation for future developments in autonomous manipulation of deformable objects.

\bibliographystyle{IEEEtran.bst}
\bibliography{references}

\vfill

\end{document}